\documentclass[a4paper,fleqn]{cas-sc}

\usepackage[authoryear,longnamesfirst]{natbib}
\usepackage{algorithm2e}
\usepackage{bm}
\usepackage{subcaption}
\usepackage{graphicx}
\RestyleAlgo{ruled}
\newtheorem{remark}{Remark}
\newtheorem{assumption}
{Assumption}
\def\tsc#1{\csdef{#1}{\textsc{\lowercase{#1}}\xspace}}
\tsc{WGM}
\tsc{QE}
\tsc{EP}
\tsc{PMS}
\tsc{BEC}
\tsc{DE}

\begin{document}
\let\WriteBookmarks\relax
\def\floatpagepagefraction{1}
\def\textpagefraction{.001}

\shorttitle{Privacy-Preserving Data Fusion for Traffic State Estimation: A Vertical Federated Learning Approach}

\shortauthors{Wang and Yang}

\title [mode = title]{Privacy-Preserving Data Fusion for Traffic State Estimation: A Vertical Federated Learning Approach}                      



%
\author[1]{Qiqing Wang}[orcid=0009-0003-0053-5910]

\ead{qiqing.wang@u.nus.edu}



\author[1]{Kaidi Yang}[orcid=0000-0001-5120-2866]
\ead{kaidi.yang@nus.edu.sg}


\cormark[1]

\affiliation[1]{organization={Department of Civil and Environmental Engineering, National University of Singapore},
    addressline={1 Engineering Drive 2}, 
    city={Singapore},
    postcode={117576}, 
    country={Singapore}}

\cortext[cor1]{Corresponding author}



\begin{abstract}
This paper proposes a privacy-preserving data fusion method for traffic state estimation (TSE). Unlike existing works that assume all data sources to be accessible by a single trusted party, we explicitly address data privacy concerns that arise in the collaboration and data sharing between multiple data owners, such as municipal authorities (MAs) and mobility providers (MPs). To this end, we propose a novel vertical federated learning (FL) approach, FedTSE, that enables multiple data owners to collaboratively train and apply a TSE model without having to exchange their private data. To enhance the applicability of the proposed FedTSE in common TSE scenarios with limited availability of ground-truth data, we further propose a privacy-preserving physics-informed FL approach, i.e., FedTSE-PI, that integrates traffic models into FL. Real-world data validation shows that the proposed methods can protect privacy while yielding similar accuracy to the oracle method without privacy considerations. 
\end{abstract}




\maketitle

\section{Introduction}
Big data applications have been receiving a significant amount of research attention in various domains of intelligent transportation systems (ITS), including traffic state estimation/prediction \citep{zheng2017estimating, herrera2010incorporation, xu2020ge, zhao2019t, zheng2020gman}, traffic control \citep{guo2019urban, liu2022optimal, amini2017big, ning2020joint}, and travel route planning \citep{huang2020multi}. Traffic big data can be generally divided into two categories. The first type is Eulerian observations measured with fixed roadside sensors (e.g., loop detectors, traffic cameras, radars, etc.) installed on a small set of road links. This type of data is typically owned by municipal authorities (MAs). The second type is Lagrangian observations obtained from vehicle trajectory data (e.g., real-time vehicle locations and speeds), typically possessed by mobility providers (MPs) with large fleets, such as ride-hailing companies and public transport operators. Although these data can be useful for ITS applications, they provide only an incomplete picture of transportation systems due to the partial spatial coverage of the road network (in terms of Eulerian observations) and partial penetration rate over the vehicle population (in terms of Lagrangian observations). Traffic state estimation (TSE) is an important research topic that leverages these partially observed traffic data to infer key traffic state variables (e.g., flow, density, speed) on road segments. Researchers have proposed a number of TSE methods based on traffic flow theory and machine learning \citep{caceres2012traffic, ke2018real, xu2020ge, fedorov2019traffic, cai2019differential,rostami2020state,saeedmanesh2021extended,van2018macroscopic,wang2016efficient,yang2018queue,zhao2019various,zhan2020link,li2021multi,wang2022real}. However, most of these methods focus on individual data sources, making them sensitive to the spatial network coverage and penetration rate. 

In order for a better estimation of traffic states, a large body of works attempt to improve the performance of TSE algorithms by combining multiple data sources, with a particular focus on the fusion between loop detector data provided by MAs and vehicle trajectories provided MPs~\citep{wang2016efficient, van2018macroscopic, wu2015cellpath,wang2018efficient,shahrbabaki2018data,ambuhl2016data, makridis2023adaptive, saeedmanesh2021extended}.
One underlying assumption behind these works is the  collaboration established between multiple data owners (i.e., MA and MPs in our context). 
Although the benefits of such collaboration have been demonstrated by game-theoretical analysis~\citep{liu2022efficient} and practical projects~\citep[e.g., the collaboration between DiDi and local authorities in several Chinese cities as presented in][]{han2019urban, zheng2018traffic}, existing studies generally focus on the operational efficiency and overlook privacy concerns.
In fact, MPs may be reluctant to contribute their data to another party since 
data sharing may leak sensitive information about their customers' mobility patterns and system operations. It has been demonstrated that human mobility patterns are highly unique and can be used to infer travelers' identities, personal profiles, and social relationships with high accuracy even when anonymization is applied \citep{de2013unique}. Customers' information has been protected by increasingly strict privacy protection regulations, e.g., the General Data Protection Regulation (GDPR) of the European Union, the violation of which can lead to huge fine. Moreover, MPs' data, even aggregated and anonymized, can provide sensitive operational characteristics such as service coverage, fleet composition, and key algorithm parameters \citep{he2019optimal}, and the leakage of such information can compromise their competitive advantages. On the other hand, MA may also be concerned that sharing traffic detector data can enable adversaries to infer signal timing algorithms or the locations of secret government facilities \citep{StravaPrivacy}, which can lead to security risks. Therefore, a key to the successful implementation of data fusion algorithms in transportation is to protect the privacy of data owners, including both MA and MPs. 

To this end, a small portion of research leverages advanced privacy-preserving mechanisms to process traffic data \citep{kim2022privacy, jin2022survey, gati2021differentially, ying2022privacysignal, liu2020privacy, parameswarath2022user, gao2022privacy}, including differential privacy (DP), secure multi-party computation (MPC), and federated learning (FL). Particularly, FL is specially designed to train data-driven models for distributed data owners, each with private data. With FL, each data owner can privately train local models with its data on cell phones or local servers and send only a portion of parameters and transformed data (instead of their raw data) to a host server for aggregation into a global model. 
The advantage of FL is that it does not require the exchange of sensitive data and can be readily integrated with state-of-the-art machine learning and optimization-based models. 
Thanks to the advantages, FL has been receiving increasing attention in various application domains, e.g., smart grid~\citep{su2021secure}, finance~\citep{antunes2022federated}, and healthcare \citep{imteaj2022leveraging}. A significant amount of works have been devoted to investigating the tradeoff between model performance~\citep{nilsson2018performance,lai2022fedscale}, privacy protection~\citep{wei2020federated,li2021survey,ma2020safeguarding}, communication efficiency~\citep{chen2021communication,hamer2020fedboost,rothchild2020fetchsgd}, computation requirements~\citep{hanzely2020lower,xia2021survey,diao2020heterofl}, and verifiability against malicious actors~\citep{chen2020zero,nguyen2023preserving}. Despite the success and of FL, the application in transportation is sparse. To the best of our knowledge, only a few works have implemented FL in the transportation context, e.g., human mobility pattern prediction using trajectory data from cell phones  \citep{feng2020pmf}, parking space estimation and traffic flow prediction algorithms from spatially distributed sensor data \citep{huang2021fedparking, liu2020privacy, xia2022short}. 

However, existing FL frameworks in transportation suffer from two limitations. First, existing frameworks are generally based on \emph{horizontal FL}, which assumes the datasets of all parties to be homogeneous in that they share the same features and structures. However, since traffic data can be extremely diverse, it is often required to perform data fusion from multiple heterogeneous datasets with different features (e.g., sensor measurement and vehicle trajectories). 
Moreover, ground-truth labels are typically possessed by MA instead of MPs since acquiring such data is typical via expensive aerial imaging, which requires special permissions from the authorities. Hence, \emph{vertical FL}, in which each data owner possesses different features and labels, can be more suitable for TSE. To the best of our knowledge, there is no work on applying vertical FL-based data fusion methods for TSE. 
Second, existing works assume that the ground-truth labels are abundant, which is unfortunately not true for TSE, as collecting ground-truth traffic states requires expensive efforts such as drone surveillance \citep{barmpounakis2020new}. Hence, existing ground-truth datasets (e.g., vehicle trajectory datasets) consist of data of only a few days, which may not be sufficient for training high-quality FL models. 

In this paper, we devise an FL-based privacy-preserving data fusion approach for MA and MPs to collaboratively develop TSE models. The contribution is two-fold. First, we leverage the promising framework of FL that enables multiple parties to collaboratively train a model without exchanging private data. Unlike existing works in transportation that focus on horizontal FL over many homogeneous edge devices (e.g., detectors, vehicles, travelers, etc.), we formulate the traffic state estimation problem as a vertical FL problem and build our algorithm on the recently developed FedBCD framework \citep{liu2022fedbcd}, which reduces communication overhead through local gradient updates and is easy to integrate with neural networks.
Second, we propose a physics-informed FL approach that integrates traffic models with FL to improve data efficiency, which ensures the applicability of the proposed FedTSE in common TSE scenarios with limited ground-truth availability. Physics-informed deep learning integrates physical models into learning-based approaches to improve the data efficiency in the training process and/or to preserve desired physical properties of the trained models, which has recently attracted increasing attention in the transportation research community~\citep{han2022physics,mo2021physics,di2023physics,lu2023physics}. However, the integration of traffic models into FL differs from classical physics-informed deep learning in that data privacy of both features and (partial) labels need to be protected in the training process. We address this challenge by combining FL with secure functional encryption. 

The rest of the paper is organized as follows. Section \ref{sec-ps} introduces the general framework for the TSE problem. Section~\ref{sec-fedtse} introduces the FedTSE approach assuming the availability of ground-truth labels. Section~\ref{sec-fedtse-case} presents the case studies and results for FedTSE. Section~\ref{sec-fedtse-pi} devises a privacy-preserving physics-informed FedTSE approach to integrate FedTSE with traffic models to improve data efficiency in training. Section~\ref{sec-fedtse-pi-case} presents the case studies and results for FedTSE-PI. Section~\ref{sec-con} concludes the paper.

\section{Problem Statement}\label{sec-ps}
In this section, we present the privacy-preserving TSE problem. Consider a typical city with an urban transportation network described by a directed graph $\mathcal{G}=(\mathcal{V},\mathcal{E})$, where each vertex $n\in \mathcal{V}$ represents a road link,  and each edge $e=(m,n)\in \mathcal{E}$ represents the connectivity from link $m\in \mathcal{V}$ to link $n\in \mathcal{V}$. Let us denote the considered time horizon as a set of discrete intervals $\mathcal{T}=\{1,2,\cdots, T\}$ of a given size $\Delta t$. 
A local MA (indexed by $k=0$) and $K$ MPs (indexed by $k=1,\cdots,K$) are interested in collaborating to develop a learning-based model to fuse their data for real-time TSE, i.e., to produce estimates $\hat{\bm{y}} = \{\hat{q}_{nt}, \hat{k}_{nt}\}_{n \in \mathcal{I},t\in\mathcal{T}}$ for flow ($\hat{q}_{nt}$) and density ($\hat{k}_{nt}$) on each road link $n\in\mathcal{V}$. 
The local MA has continuous access to traffic count measurements from loop detectors installed on a small portion of roads $\mathcal{L}\subset \mathcal{V}$. Let us denote $c_{nt}$ as the traffic count at time step $t \in \mathcal{T}$ measured by the loop detector installed on road link $n\in \mathcal{L}$. MA may also have conducted expensive aerial imaging to collect ground-truth flow and density information across the entire transportation network over the time horizon $\mathcal{T}$. Let us denote the ground-truth flow and density of road $e$ at time step $t$ as $q_{nt}$ and $k_{nt}$, respectively. 
Each local MP manages a fleet of vehicles denoted as $\mathcal{R}_k$, which provides real-time vehicle trajectory information across the network. Notice that the fleets owned by MPs collectively constitute a small portion of the entire vehicle population on the road. 
At each time step, the vehicle trajectories of each MP are characterized by a set of features, such as the origin-destination (OD) demand $\bm{\mu}_t^k = \{\mu_{odt}^k\}_{o,d\in\mathcal{V}}$, travel time $\bm{\tau}_t^k = \{\tau_{nt}^{rk}\}_{n \in \mathcal{V},r\in\mathcal{R}_k}$ and travel distance $\bm{\xi}_t^k = \{\xi_{nt}^{rk}\}_{n \in \mathcal{V},r\in\mathcal{R}_k}$ of each vehicle belonging to MP $k$ on each road link, turning ratios $\bm{\zeta}_t^k = \{\zeta_{et}^k\}_{e \in \mathcal{E}}$ observed by each MP between any adjacent pair of links, etc. All these features can provide valuable information about traffic states, e.g., the total travel time and total travel distance can serve as a proxy of density and flow, respectively. We further note that MPs can be heterogeneous in that the features used by individual MPs do not have to be the same. 

Following the previous description, we can summarize the dataset of data owner $k$ (MA or MP) as $\mathcal{D}^k=\{\bm{d}_t^k\}_{t\in\mathcal{T}}$, which consists of $T$ data samples, and each sample $\bm{d}_t^k$ corresponds to a time step $t$. Specifically, $\bm{d}_t^k$ can be written as
\begin{align}\label{eq9}
    \bm{d}_t^k = \left\{
    \begin{aligned}
    &(\bm{x}_t^k, \bm{y}_t),\quad & \mathrm{if }~k=0,~t\in\mathcal{T}\\
    &\bm{x}_t^k,\quad & \mathrm{if }~k=1,\cdots,K,~t\in\mathcal{T}
    \end{aligned}\right.
\end{align}
where $\bm{y}_t = \left\{q_{nt},k_{nt}\right\}_{n\in \mathcal{V}}$ represents the ground-truth label at time step $t\in\mathcal{T}$, and $\bm{x}_t^k$ summarizes the features corresponding to time step $t$ of data owners $k$, represented as
\begin{align}
    \bm{x}_t^k = \left\{
    \begin{aligned}
    &\left\{c_{et}\right\}_{e\in \mathcal{L}},\quad & \mathrm{if }~k=0\\
    &\left\{\bm{\mu}_t^k, \bm{\tau}_t^k, \bm{\xi}_t^k, \bm{\zeta}_t^k, ...\right\}_{i \in \mathcal{E}},\quad & \mathrm{if }~k=1,\cdots,K
    \end{aligned}\right.
\end{align}

The data of MPs can be beneficial for MA to understand the traffic conditions in the transportation network, especially the links on which MA does not have detectors. However, MPs can be reluctant to share their data due to the fear of privacy leakage. Even if MPs only share aggregated data with MA, such data can still be used to infer sensitive information about MPs' operations, such as their service coverage, algorithm specifics, etc., which could potentially hinder their competitive advantages. Hence, our main goal is to facilitate collaboration while preserving the data privacy of both MA and MPs.

\begin{figure}
    \centering
    \includegraphics[width=0.8\textwidth]{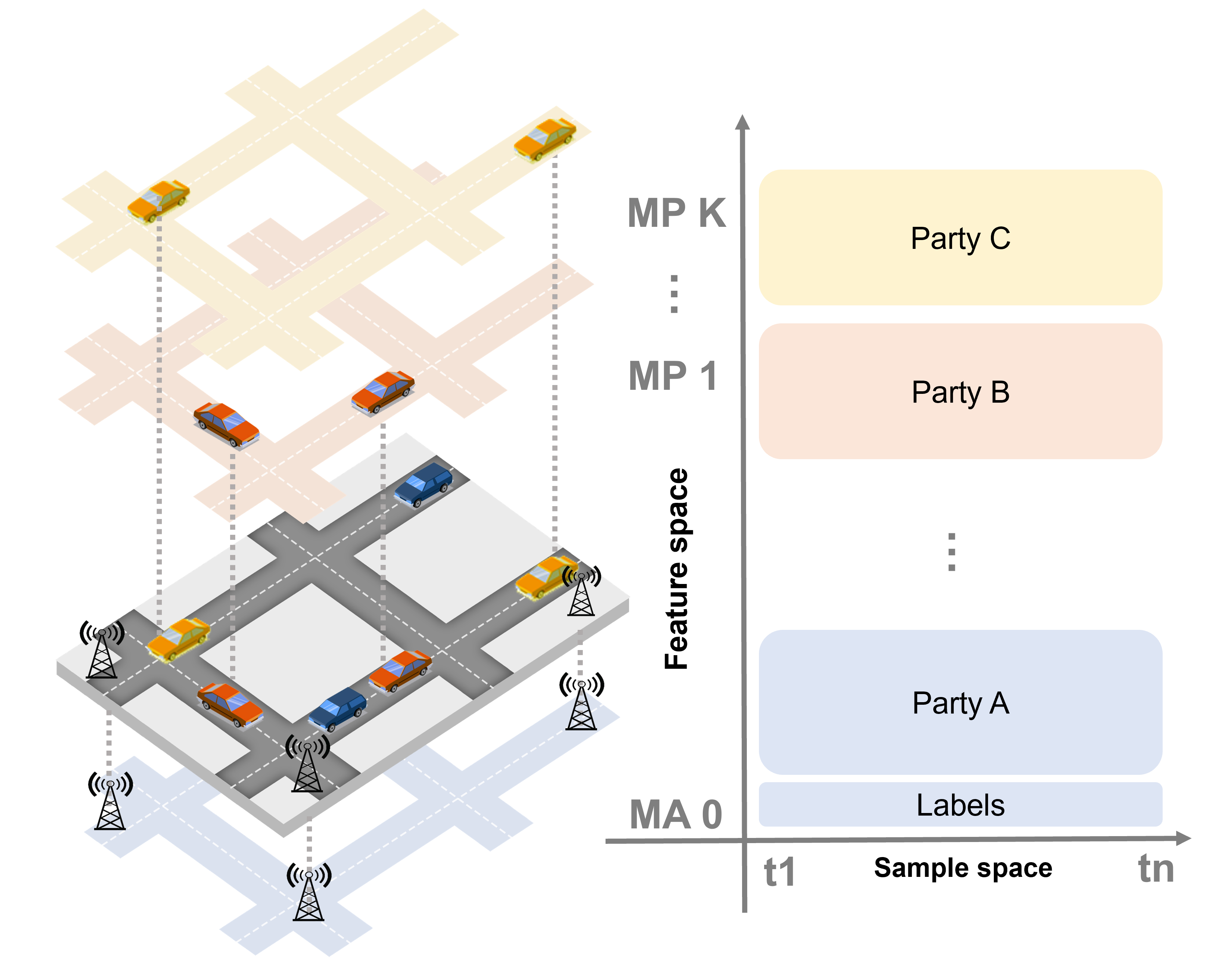}
    \caption{Vertically partitioned traffic data in urban transportation network}
    \label{fig:data}
\end{figure}

We would like to highlight the key difference between our problem setting and that of existing FL-based privacy-preserving works in traffic systems such as \cite{liu2020privacy}. Existing works leverage horizontal FL in edge computing scenarios where data is horizontally distributed over edge devices, e.g., sensors for specific traffic regions or road segments, and the data at different devices typically contains identical features due to homogeneous sensors. These works aim to enable the edge nodes to jointly train a learning-based model (typically identical for each edge device) without these devices having to share data. Although privacy can be protected with these methods, their main benefits lie in the reduction of communication costs. In our problem, however, data is vertically distributed within multiple data owners with separate measurements over the urban network, which can be seen as different features such as loop detector data and trajectory information, as shown in Figure~\ref{fig:data}. In this type of problem, privacy can be an important consideration because data can contain sensitive trade secrets of each data owner. 
To the best of our knowledge, FL for such problems with vertical data segmentation has rarely been explored in transportation literature. Therefore, we devise a vertical FL-based approach to perform TSE with privacy-preserving fusing of MA and MPs' data. 

Before proceeding to introduce our vertical FL-based framework, let us summarize the following assumptions we make about MA and MPs. 

First, our problem settings implicitly impose an assumption that MPs are willing to collaborate with MA by contributing data (in a privacy-preserving manner) and computation resources for MA to perform TSE, as presented in 
Assumption~\ref{asm:computation_MP}.  

\begin{assumption}[Willingness of MPs to collaborate] \label{asm:computation_MP}
    MPs are willing to contribute their data and computation resources for the training and deployment of the proposed FedTSE algorithms. 
\end{assumption}

\begin{remark}[Willingness of MPs to collaborate]
    We make the following remarks for Assumption~\ref{asm:computation_MP}. First,  contributing to MA's traffic state estimation and control can help alleviate traffic congestion, which in turn reduces the travel time of MPs' fleets and hence benefits MPs' operations. More broadly, the willingness of data sharing of data owners can be modeled as a cooperative game \citep{jia2019towards,donahue2021model} or coopetitive game \citep{liu2022efficient}, and incentives have been developed to encourage the participation of data owners, especially in FL settings \citep{jiang2022vf,zeng2021comprehensive}. 
Second, in terms of computation resources, it is natural for MPs, such as ridesharing companies and mapping companies, to be equipped with sufficient computational resources since their own operations require solving complex optimization problems. Hence, they have the capability to perform internal computation, as required by FL. 
Third, it is worth noting that the collaboration between MPs and MA not only resides in scientific research but also has been implemented in practice. For example, DiDi, a ridesharing company in China, has contributed its trajectory data to the local authorities in several cities to improve signal timing and ramp metering algorithms~\citep{han2019urban, zheng2018traffic}, which sheds light on the practical significance of the proposed framework.
\end{remark}
 
Then, we make the following assumption on the data-sharing behavior of MA and MPs, as described in Assumption~\ref{asm:honest-but-curious}. 
\begin{assumption}[Honest-but-curious MA and MPs] \label{asm:honest-but-curious}
    We assume that both types of data owners, i.e., MA and MPs, are honest but curious, meaning that they will strictly follow the communication and computation protocols with no intention to modify their data to achieve better benefits, but may attempt to infer the data of other data owners. 
\end{assumption}

Assumption~\ref{asm:honest-but-curious} is a commonly-made assumption in privacy-related literature ~\citep{TsaoYangGopalakrishnanPavone2022,tan2024privacy}. We make the following remarks on the implications of this assumption in our context. The first remark is about curious MA and MPs, as well as their privacy concerns. 
\begin{remark}[Curious MA/MPs and sensitive information] We consider both MA and MPs to be curious, as MPs are naturally interested in learning the operations of other MPs due to competition, and adversaries may gain unauthorized access to the transmitted data to MA. Specifically, we consider the following information of MA and MPs to be sensitive. First, MPs are interested in protecting the information about their customers and business operations, including passenger origin-destination pairs (ODs) and vehicle trajectories that reveal passenger information, as well as the aggregated trajectories on the traversed links that can be used to infer the service area, demand patterns, fleet size, and algorithm parameters \citep{he2020optimal}, both of which are essential for MPs to maintain their competitive advantage.   
Second, MA is mainly concerned that the leakage of their data can enable adversaries to infer traffic operational algorithms (e.g., signal timing algorithms) and locations of sensitive government facilities, which might be leveraged by adversaries to perform attacks.   
\end{remark}

Further note that in Assumption~\ref{asm:honest-but-curious}, we consider honest MPs to simplify the discussion. In practice, MPs may be incentivized to send forged data if they can achieve higher benefits, especially with privacy-preserving mechanisms that can make it hard to verify MPs' data. In such cases, we can relax this assumption by incorporating a verification mechanism based on zero-knowledge proof \citep{fiege1987zero} to prevent MPs' strategic behavior, as described in Remark~\ref{rmk:zkp}.

\begin{remark}[Verification of MP data via zero-knowledge proof] \label{rmk:zkp}
 Zero-knowledge proof enables a prover (i.e., MP in our context) to prove to a verifier (i.e., MA in our context) that it is performing the correct calculations with the true data without explicitly sharing the data. Our previous work \citep{tsao2022trust} proposed a zero-knowledge proof-based data-sharing protocol between MA and MPs, such that MA can outsource their calculations to MPs and be able to verify the calculation results without requiring MPs to share their true data. Such calculations can include finding the value of a function, checking if some conditions are met, solving a convex optimization problem, etc. Naturally, since MPs in the FL framework mainly perform simple calculations such as gradient steps and forward propagation (see Section~\ref{sec-fedtse}),  FL and zero-knowledge proof can be combined to improve the verifiability of the information shared by MPs to prevent strategic behavior \citep{chen2020zero,nguyen2023preserving,ghodsi2023zprobe,li2021privacy}, which is, however, beyond the scope of the paper. 
    
\end{remark}
 
With the problem settings and assumptions, we now proceed to introduce our proposed FL-based privacy-preserving data fusion framework for TSE. 

\section{FedTSE: Vertical Federated Traffic State Estimation}\label{sec-fedtse}
\begin{figure}[!htbp]
	\centering
    \includegraphics[width=0.9\textwidth]{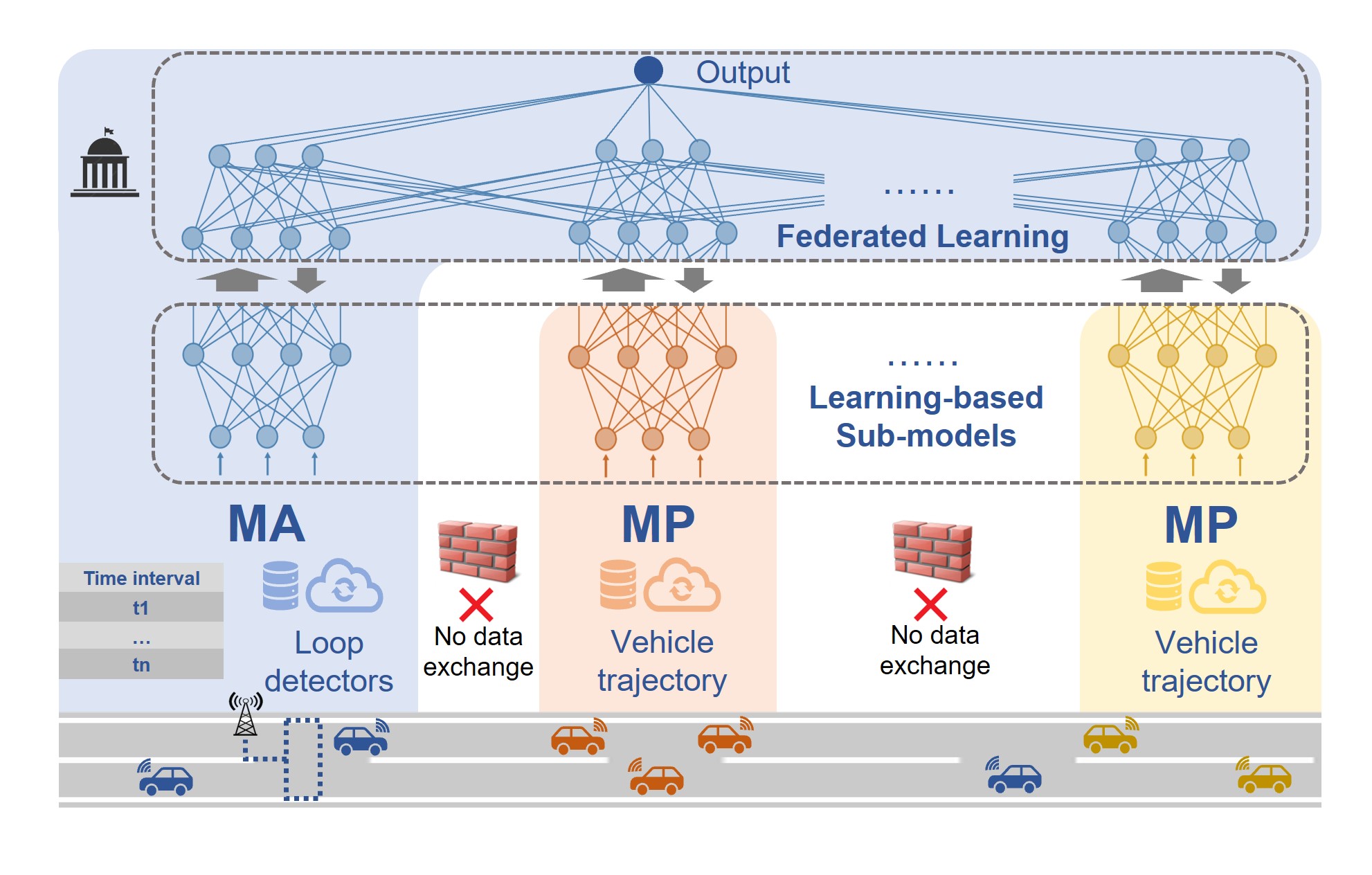}
	\caption{Schematic illustration of FedTSE, where MA (in blue) and multiple MPs (in orange and yellow) jointly train a neural network for TSE while keeping their data private. Only the values and gradients of private sub-models are exchanged between parties. }
	\label{fig1}
\end{figure}

In this section, we present a vertical FL-based data fusion approach, hereafter named FedTSE, to enable MA to improve TSE accuracy with MPs' data while protecting the data privacy of all data owners. The framework of FedTSE is illustrated in Figure~\ref{fig1}, where MA serves as the host (in blue) and each MP serves as a guest (in red and yellow). 

In this framework, each data owner $k$ (MP or MA) maintains a private sub-model $\bm{\phi}^k(\cdot)$ only accessible by the data owner itself, which can be represented as a neural network parameterized by parameters $\bm{\theta}^k$.  These models are used and trained in a collaborative manner using private datasets $\mathcal{D}^k$ that will always stay with their owners. 

As shown in Figure~\ref{fig1}, at each time step, MP $k$ processes its data $\bm{x}^k$ according to a sub-model $\bm{z}^k=\bm{\phi}^k(\bm{x}^k;\bm{\theta}^k)$ with $\bm{\theta}^k$  as private parameters to be determined. The sub-model output $\bm{z}^k$ will be sent to MA.  MA, after collecting all sub-model output $\bm{z}^k$ from each MP, will perform TSE according to its sub-model $\bm{y}=\bm{\phi}^0(\bm{x}^0, \bm{z}^1, \cdots, \bm{z}^k; \bm{\theta}^0)$ with private parameters $\bm{\theta}^0$ to be determined.  

In order to obtain the parameters $\Theta=\left(\bm{\theta}^0, \cdots,  \bm{\theta}^k\right)$ for all data owners, the FL framework aims to solve a collaborative training problem written as: 
\begin{align}\label{311}
	\min _{\Theta} L(\Theta ; \mathcal{D}) \triangleq \frac{1}{T} \sum_{t=1}^{T} f\left(\bm{\theta}^{0}, \ldots, \bm{\theta}^{K} ; \bm{d}_t^0,\cdots, \bm{d}_t^K\right)+\lambda \sum_{k=0}^{K} \gamma\left(\bm{\theta}^{k}\right)
\end{align}
where the loss function $f(\cdot)$ and regularizer $\gamma(\cdot)$ are combined with a weight parameter $\lambda$. The loss function can be represented in the form of  $f\left(\bm{\phi}^0(\bm{x}^0_t, \bm{z}^1_t, \cdots, \bm{z}^k_t; \bm{\theta}^0), \bm{y}_t\right)$ to penalize the deviation between estimated states $\bm{\phi}^0(\bm{x}^0_t, \bm{z}^1_t, \cdots, \bm{z}^k_t; \bm{\theta}^0)$ and true states $\bm{y}_t$, where  $\bm{z}^k_t=\bm{\phi}^k(\bm{x}^k_t;\bm{\theta}^k)$ is the output from the private model of MP $k$ at time step $t$. 
Specifically, we leverage the following loss function for TSE to penalize the mean squared errors of flow and density estimation:  
\begin{align}\label{312}
    f\left(\Theta ; \mathcal{D} \right) = f_k\left(\Theta ; \mathcal{D} \right) + 
 f_q\left(\Theta ; \mathcal{D} \right) 
     =  \frac{1}{N} \sum_{v \in \mathcal{V}} \left\|k_{vt}-\hat k_{vt}\right\|^2 + \alpha \frac{1}{N} \sum_{v \in \mathcal{V}} \left\|q_{vt}-\hat q_{vt}\right\|^2 
\end{align}
where $\alpha$ represents the weighing factor between the two terms. 

Figure \ref{311} can be jointly solved by MA and MPs using variants of the Federated Stochastic Gradient Descent (Fed-SGD) approach \citep{mcmahan2017communication, liu2022fedbcd} without explicitly sharing their raw data. We employ a Federated Stochastic Block Coordinate Descent (FedBCD) \citep{liu2022fedbcd} approach that enables a sufficient number of local updates to account for expensive communication overhead in FL training.

\begin{algorithm}[hbt!]
\caption{Algorithm for FedTSE}\label{alg:1}
\DontPrintSemicolon
\SetAlgoLined
\textbf{Input:} Private datasets $\mathcal{D}^k$ and learning rate $\eta_k$ for data owner $k = 0, 1, \ldots, K$, regularizer weight parameter $\lambda$, and the number of local updates $Q$\\
\textbf{Output:} $\Theta = (\bm{\theta}^0, \bm{\theta}^1, \ldots, \bm{\theta}^k)$\\

\For{each iteration $i = 1, 2, \ldots$}{
    \If{ $i\mod Q = 0$}{
        MA randomly samples a mini-batch of time steps $\mathcal{B} \in \mathcal{T}$ and synchronizes it with MPs\;
        \For{$k = 1, \ldots, K$ in parallel}{
            MP $k$ computes local output $\bm{z}^k_t = \bm{\phi}^k(\bm{x}^k_t; \bm{\theta}^k)$ with its private sub model for $t \in \mathcal{B}$\;
            MP $k$ sends ${\bm{z}^k_t}$ to MA\;
        }
        
        MA computes gradients $\frac{\partial L}{\partial \bm{\theta}^0}$ and updates sub-model parameters according to $\bm{\theta}_0^{i+1}=\bm{\theta}_0^i-\eta_0 \frac{\partial L}{\partial \bm{\theta}^0} $\;
        MA computes and sends gradients $\frac{\partial L}{\partial \bm{z}^k_t}$ to each MP $k$\;
    }

    \For{$k = 1, \ldots, K$ in parallel}{
        MP $k$ computes $\frac{\partial L}{\partial \bm{\theta}^k}$ with the most recent $\frac{\partial L}{\partial \bm{z}^k_t}$ according to Equation \ref{gradient}\;
        MP $k$ Update $\bm{\theta}_k^{i+1}=\bm{\theta}_k^i-\eta_k \frac{\partial L}{\partial \bm{\theta}^k}$\;
    }
    \If{convergence criterion met}{
        break\;
    }
    }
\end{algorithm}

Algorithm \ref{alg:1} describes the training procedure of FedTSE, which requires real-time communication between MA and MPs. Within each communication round, MA samples a mini-batch of time steps $\mathcal{B} \in \mathcal{T}$ and shares it with MPs to synchronize the data used for training.
Each MP then computes its local model output $\bm{z}^k_t=\bm{\phi}^k(\bm{x}^k_t; \bm{\theta}^k)$ (i.e., intermediate results) and sends it to MA. 
After receiving the sub-model output $\bm{z}^k_t$ from all MPs, MA computes two gradients:
(1) the gradient of $L$ with respect to the model parameters of MA ($\bm{\theta}^0$), and 
(2) the gradients of $L$ with respect to the output $\bm{z}_t^k$ of each MP, i.e., $\frac{\partial L}{\partial \bm{z}^k_t}$, which will be sent back to MP $k$. 
MP $k$ uses the received gradient $\frac{\partial L}{\partial \bm{z}^k_t}$ to calculate the gradient of the loss function $L$ with respect to its sub-model parameters $\theta_k$ based on the chain rule of partial derivatives shown in  by  Equation \ref{gradient}: 
\begin{align}\label{gradient}
    \frac{\partial L}{\partial \bm{\theta}^k}=\sum_{t \in \mathcal{B}} \frac{\partial L}{\partial \bm{z}^k_t} \frac{\partial \bm{z}^k_t}{\partial \bm{\theta}^k},
\end{align}
which will be used by MP $k$ to update its sub-model parameter $\bm{\theta}^k$. Notice that since the sub-models are private, MA does not have any information about $\frac{\partial \bm{z}^k_t}{\partial \bm{\theta}^k}$ and hence $\frac{\partial L}{\partial \bm{\theta}^k}$. 

After data owners communicate and derive the relevant gradients, they will use these gradients to update their sub-model parameters according to a gradient-descent rule $\bm{\theta}^{k,i+1}=\bm{\theta}^{k,i}-\eta_k \frac{\partial L}{\partial \bm{\theta}^k} $ with $\eta_k$ indicating the learning rate of data owner $k$. 
It is worth noting that the sharing of relevant gradients and sub-model output incurs expensive communication overhead between MA and MP \citep{liu2022vertical, wei2022vertical}. We adopt a straightforward method to reduce the amount of data exchanged between MA and MPs by allowing them to locally update the submodel for $Q>1$ rounds without exchanging information, using the previously received gradient information. Such local updates have been widely used in existing literature, and it has been proved that the communication overhead can be significantly reduced without significantly impacting FL performance if $Q$ is chosen appropriately \citep{liu2022fedbcd}. 

We make the following remarks to better discuss the properties of FedTSE. 
As the main motivation for adopting FL is privacy, we make Remark~\ref{rmk:privacy} to discuss the privacy guarantee provided by FedTSE for each data owner.
\begin{remark}[Privacy of data owners in FedTSE] \label{rmk:privacy}
    Similar to \citet{liu2022fedbcd}, let us say the data privacy of data owner $k$ (MP or MA) is preserved if one cannot uniquely determine its true data $\bm{x}^k$ from its exchanged messages, either the sub-model output $\bm{z}^k$ shared by MP $k$ to MA and/or the corresponding gradients $\frac{\partial L}{\partial \bm{z}^k}$ sent from MA to MPs. 
    With such a privacy notion, the developed FedTSE algorithm can protect the data privacy of all data owners. The reason is two-fold. First, the sub-models owned by individual MPs and the MA are private, meaning that the data of both types of data owners are transformed via an unknown transformation.  \citet{liu2022fedbcd} has proved that as long as the feature dimension of data owner $k$ is no less than 2, the exchanged message from data owner $k$ corresponds to infinite possibilities of true states $\bm{x}^k$. This is rather intuitive since we can  tweak the true states $\bm{x}^k$ and the sub-model $\bm{\phi}^k(\cdot)$ in a joint manner without changing the sub-model output $\bm{z}^k$ for MP and the exchanged gradients $\frac{\partial L}{\partial \bm{z}^k}$ for MA.  Second, we reduce the dimension of the sub-model output to be significantly lower than that of features. This further helps to reduce the information shared between MA and MPs, and thus strengthen privacy protection. 
    Further note that such a privacy guarantee can be enhanced by integrating with other privacy-preserving mechanisms, such as differential privacy \citep{dwork2006differential,TsaoYangGopalakrishnanPavone2022} and secure multiparty computation \citep{Shamir79,tan2024privacy}, which will serve as future work. 
\end{remark}

Remark~\ref{rmk:computation} shows that FedTSE does not impose significant computational burdens on MPs.  
\begin{remark}[Computation requirements of FedTSE] \label{rmk:computation}
    FedTSE does not impose heavy computation burden to MP. In particular, each MP's private model, i.e., the local neural network, is relatively light-weighted with an input dimension of $H|\mathcal{V}|$ and an output dimension of no greater than $|\mathcal{V}|$, where $H$ represents the size of feature space for each edge (with a typical value of 20 in our case studies), and $|\mathcal{V}|$ represents the number of edges. Hence, the training and the deployment of such neural networks would be relatively affordable for MPs. In fact, for our case study, the neural networks can be trained efficiently on a PC with NVIDIA GeForce RTX 2060 within 30 minutes and on a server with NVIDIA A100 within 10 minutes.  
\end{remark}

\section{Case study on FedTSE}\label{sec-fedtse-case}
\subsection{Settings of the case study} \label{sec:setting_fedtse}
\subsubsection{Datasets}
In this paper, we evaluate the performance of FedTSE by performing a case study of a signalized urban corridor in Athens, Greece. The case study encompasses both a real-world dataset,  pNEUMA \citep{barmpounakis2020new}, and a simulated dataset generated from a well-calibrated  simulation created by SUMO \citep{lopez2018microscopic} for this site. The reason for involving simulated data is because the real-world data is limited and insufficient for comprehensive analysis. Consequently, the real-world data is used to evaluate the performance of FedTSE, while the simulated data is used to conduct comprehensive sensitivity analysis. 

\begin{figure}[!htbp]
    \centering
    \includegraphics[width=1\textwidth]{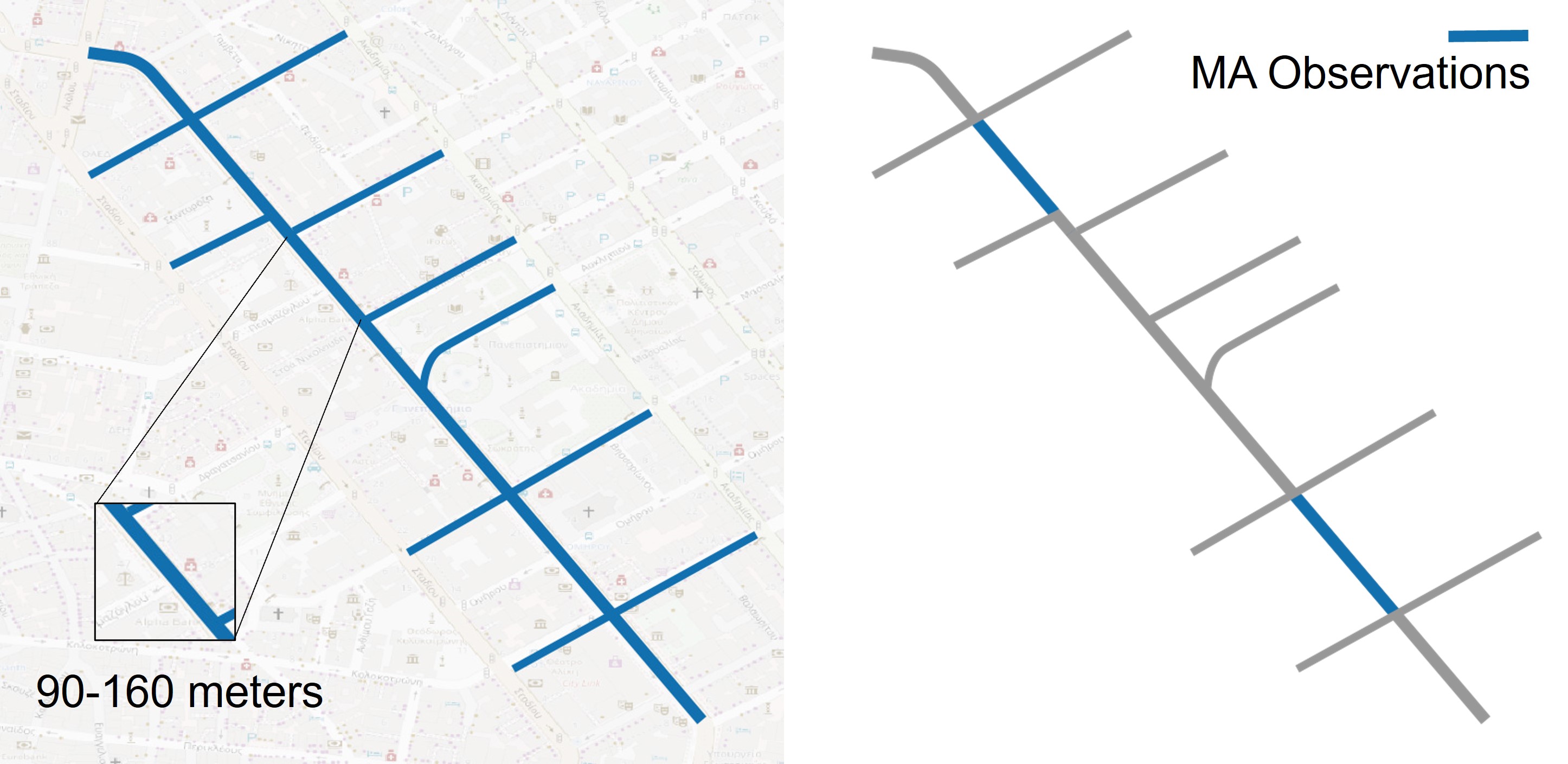}
    \caption{pNEUMA urban road corridor}
    \label{fig:pneuma-map}
\end{figure}

\textit{Real-world dataset:} The real-world dataset pNEUMA is adopted for the evaluation of FedTSE, where a fleet of ten drones recorded trajectories of vehicles operating in the central district of Athens, Greece, via drone imaging. The trajectories covered the morning peak on four weekdays (i.e. 24/10/2018, 29/10/2018, 30/10/2018 and 01/11/2018), from 8:30 to 10:30 a.m. Due to the battery capacity limitations of the drones, the trajectories were recorded at 30-minute intervals by the swarm in sequential sessions with blind gaps (i.e. effective recording interval is less than 30 minutes). A main road corridor named Panepistimiou Street, a primary road that runs one way for non-transit vehicles, together with its side links, is selected as the studied site. This corridor consists of 7 intersections with fixed-time signal timing strategies and 17 links, each with 90-160 meters in length. The geometric layout of the corridor and the locations of the detectors are shown in Figure \ref{fig:pneuma-map}, whereby we generally assume that only two links are equipped with MA's loop detectors. We associate vehicle trajectories in the pNEUMA dataset to links using open-source map-matching software called LeuvenMapMatching \citep{meert2018hmm}. The goal of TSE is to estimate the traffic states of 9 links with relatively high traffic flow.   

\textit{Simulation dataset:} Due to the limited amount of real-world data, we built a SUMO simulation to provide a more comprehensive analysis. We use the real-world trajectory data of the four weekdays collected from pNEUMA to calibrate our simulation, including the traffic signals, demand, and car following models. Virtual traffic detectors were created to collect the traffic flow data. Our traffic simulation was conducted over 810 minutes, with the initial 10 minutes serving as a warm-up phase for the simulation to ensure its stability and accuracy. The subsequent 800 minutes of data collection were then utilized to evaluate our estimations. We also assume that only two links in this main road corridor are equipped with MA's loop detectors as shown in Figure \ref{fig:pneuma-map}.

\subsubsection{Benchmarks for FedTSE}
To evaluate the performance of FedTSE, we compare the following benchmarks. Recall that here MA is assumed to have access to ground-truth traffic states, and hence data fusion for FedTSE refers to the expansion of features with MPs' data. Here, we consider the scenario with one MP. 

\begin{itemize}
    \item \textbf{TSE-n (no data fusion)}: This benchmark corresponds to the scenario where MP is not willing to share any data with MA due to privacy concerns, and MA estimates traffic states only using its own data from a limited number of fixed observers in set $\mathcal{L}$ (i.e., two observations as shown in Figure \ref{fig:pneuma-map}) without fusing with MPs' features. Specifically, MA's feature at each time step $t$ is selected as the past $h$-step loop detector data of all links with detection, i.e., $\{c_{nt'}\}_{n\in\mathcal{L}, t' \in \{t-h+1,\cdots,t\}}$.     

    \item \textbf{TSE-p (partial data fusion without privacy protection)}: This benchmark algorithm corresponds to the scenario where MP directly shares the real-time speed data calculated from its operating fleets without considering privacy protection, and MA integrates both its own loop detector data and MP's speed data to train a TSE model. 
    Specifically, MA's feature can be represented by the past $h$-step loop detector data, similar to TSE-n, and MP's feature can be represented by the past $h$-step speed information, represented by $\{\mu_{nt}\}_{n\in\mathcal{V}, t' \in \{t-h+1,\cdots,t\}}$ , where the speed of link $n\in\mathcal{V}$ at time s tep $t$ is represented by $\mu_{nt} = \Big(\sum_{r\in\mathcal{R}}{\xi_{nt}^{r}}\Big)/\Big(\sum_{r\in\mathcal{R}}{\tau_{nt}^{r}}\Big)$, i.e., the ratio between total travel distance and total travel time. 
    We envision that this scheme of collaboration can be accepted by some MPs since the sensitive information contained in the speed data is much truncated compared to raw trajectory data.  Nevertheless,  note that even though the speed data is highly aggregated, it may still reveal MPs' service areas. This benchmark is leveraged to demonstrate the necessity of performing more extensive data fusion with privacy considerations. 

    \item \textbf{Orcale (data fusion without privacy protection)}: This benchmark corresponds to the scenario where MP fully trusts MA and is willing to share all its trajectory data to MA without privacy protections. Therefore, MA trains FedTSE with loop detectors’ data and MP's data, including the past $h$-step total travel time and total travel distance. Specifically, MA's features are identical to those of TSE-n and TSE-p, and MP's features can be represented by $\Big\{\sum_{r\in\mathcal{R}}{\xi_{nt}^{r}}, \sum_{r\in\mathcal{R}}{\tau_{nt}^{r}}\Big\}_{n\in\mathcal{V}, t' \in \{t-h+1,\cdots,t\}}$. 
    This benchmark serves as the oracle to quantify a performance upper bound we expect to achieve via data fusion. 

    \item \textbf{Orcale-cell (extensive data fusion without privacy protection)}: This benchmark is similar to Oracle, and the only difference is that MP further divides its link to multiple cells (6 in our experiment) and a time step into multiple sub-steps (2 sub-steps in our experiment), and uses the total travel time and total travel distance within each cell and sub-step as features. Such finer representation maintains more information in the trajectory data, which has better potential in improving TSE yet worse privacy concerns if shared directly. This benchmark is used to evaluate whether higher-resolution data fusion could enhance TSE performance. 

    \item \textbf{FedTSE (Q=1)}: the proposed privacy-preserving data fusion vertical federated learning framework (FedTSE) with one local update. The features of both MA and MPs are identical to Oracle. 
    
    \item \textbf{FedTSE-cell (Q=1)}: FedTSE whereby MP provides data with cell representation. The features of both MA and MPs are identical to Oracle-cell. This benchmark is leveraged to highlight the benefits of vertical FL. Since privacy is protected, vertical FL can encourage MPs to use higher-resolution features, which could in turn improve the TSE performance.  

    \item \textbf{FedTSE (Q=2)}: FedTSE with two local updates. The features of both MA and MPs are identical to Oracle. 

    \item \textbf{FedTSE (Q=3)}: FedTSE with three local updates. The features of both MA and MPs are identical to Oracle. 

\end{itemize}

\subsubsection{Experiment settings}
We next present our experiment settings, including the experiment scenario and model architecture. 

\textit{Experiment scenario:} As mentioned above, let us consider the scenario with assume one MA and one MP collaborate to estimate traffic states (i.e., density and flow) of our studied site of each time interval of length  $\Delta t = 10$ seconds. 
MA is assumed to have access to the ground-truth data (i.e. real density and flow on each link). 
MA further provides loop detector data on two links as illustrated in Figure~\ref{fig:pneuma-map}. MP provides the features corresponding to each of the aforementioned benchmarks associated with its operating fleets at each time step $t$. For both MA and MP, we use the past $h=9$ steps of all the features to estimate the traffic states at the current time step.

\textit{Model architecture:} MA initiates two models: a global model and a sub-model. The global model (maintained by MA) has an architecture of a simple Multilayer Perceptron (MLP), and the sub-model of MA is based on the Spatio-Temporal Graph Convolutional Networks (STGCN) proposed by \cite{yu2017spatio}, which is a well-adopted learning-based model for traffic flow estimation and prediction. The private sub-model employed by MP also leverages STGCN to extract complex spatio-temporal information. In FedTSE, the input dimensions for MA and MP sub-models are $9*2*17=306$. Features that are not observed are filled with zero. The local output (i.e., sub-model output) dimensions of MA and MP are 9. The final output dimension of MA's global model is 18. This reduced dimension setting from input 306 to local output 9 is aligned with the privacy guarantee of FedTSE as discussed in Remark~\ref{rmk:privacy}. 

It's important to note that the architecture of both the global and sub-models for MA and MP can be changed as needed. Furthermore, MA and MP are designed not to have knowledge of each other's model architecture and model parameters. To ensure uniformity across different benchmarks in our case study, all benchmarks are configured with the same architecture for sub-models and global models as those in FedTSE. 
For benchmarks not utilizing the FL structure, we assume a same architecture with an STGCN followed by an MLP. This assumption is to maintain consistency in the model architecture for further comparison.

\textit{Training settings:} The learning rate for sub-models STGCN and the global model MLP is set to 3e-4. For all models, we use data from the first 80\% for training and the last 20\% for testing. 12.5\% of the training samples are selected for validation. The batch size is 128 during training. All experiments are conducted using PyTorch on a server with NVIDIA A100. We adopt root mean square error (RMSE) and mean absolute error (MAE) as the performance criteria to evaluate the benchmarks.

\subsection{Performance of FedTSE} \label{sec:fedtse_performance}
\begin{figure}[!htbp]
     \centering
     \begin{subfigure}[b]{0.49\textwidth}
         \centering
         \includegraphics[width=\textwidth]{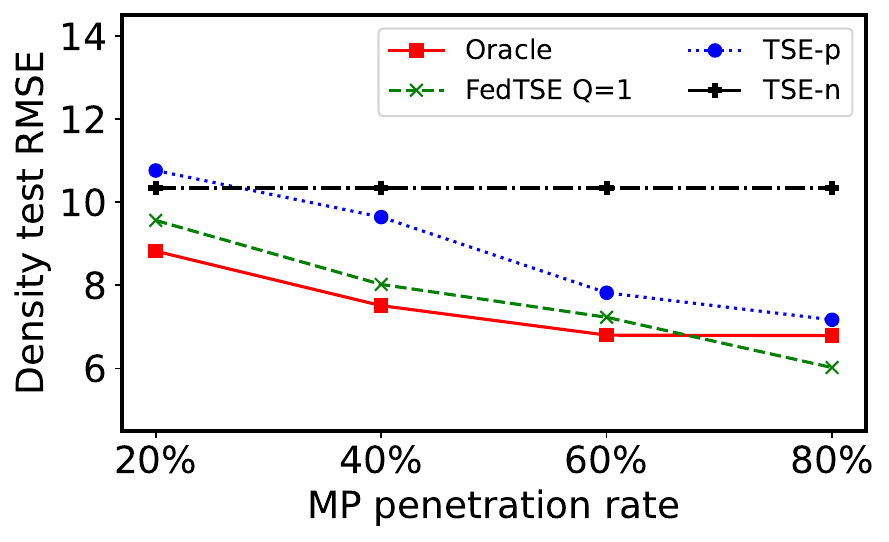}
         \caption{Density test RMSE on real-world data}
         \label{fig:pr1}
     \end{subfigure}
     \begin{subfigure}[b]{0.49\textwidth}
         \centering
         \includegraphics[width=\textwidth]{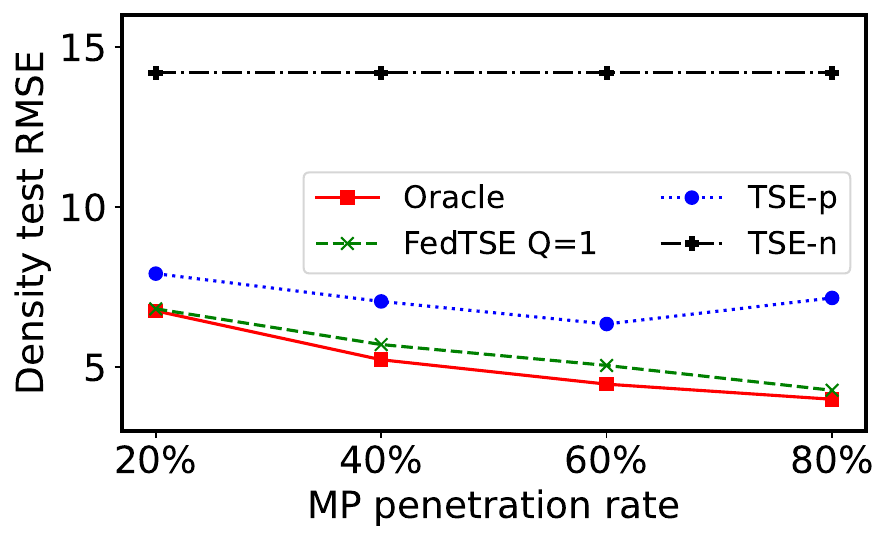}
         \caption{Density test RMSE on simulation data}
         \label{fig:pr2}
     \end{subfigure}
     \caption{Model performance comparison with different MP penetration rate.}
    \label{fig:pr}
\end{figure}

\emph{Performance on real-world data}: Table \ref{realtab1} and Figure \ref{fig:pr} compare the TSE performance of different benchmarks on real-world data. The benchmarks with the best estimation results are \underline{underlined}, while those ranked second are highlighted in \textbf{bold}. As TSE-n does not have data fusion, its error is the same for different MP penetration rates, and the performance is the worst compared to other benchmarks. 
We can see the trade-off between privacy and estimation accuracy by comparing FedTSE and Oracle, which in general have similar performance. 
This suggests that FedTSE can protect privacy with a marginal impact on the estimation performance.
It is worth noting that FedTSE outperforms TSE-n and TSE-p whereby MP shares no features or only partial features due to privacy concerns. 
This suggests that despite the trade-off between privacy and utility, FedTSE can potentially improve TSE performance by encouraging MP to share more features, which sheds light on the value of privacy protection  in data fusion. 

\begin{table}[!htbp]
	\centering
	\caption{Model performance comparison on real-world dataset.}
	\label{realtab1}
		\begin{tabular}{lccccccccc}
        \hline
                      & Metrics & \multicolumn{4}{c}{Density (veh/km/lane)}         & \multicolumn{4}{c}{Flow (veh/min/lane)}     \\
                      \hline
MP penetration rate (\%)     &         & 20    & 40    & 60    & 80    & 20   & 40   & 60   & 80   \\
\hline
\multirow{2}{*}{TSE-n}     & RMSE    & 10.34 & 10.34 & 10.34 & 10.34 & 1.91 & 1.91 & 1.91 & 1.91 \\
                        & MAE     & 7.19  & 7.19  & 7.19  & 7.19  & 1.50 & 1.50 & 1.50 & 1.50 \\
\multirow{2}{*}{TSE-p}     & RMSE    & 10.76 & 9.64  & 7.82  & 7.17  & 1.24 & 1.15 & 1.09 & 1.12 \\
                        & MAE     & 7.01  & 6.35  & 5.28  & 4.67  & 0.97 & 0.91 & 0.85 & 0.88 \\
\multirow{2}{*}{Oracle}     & RMSE    & \underline{8.82}  & \underline{7.51}  & \underline{6.80}  & \textbf{6.79}  & \textbf{1.19} & \textbf{1.11} & \textbf{1.05} & \underline{0.99} \\
                        & MAE     & \underline{5.85}  & \underline{4.83}  & \underline{4.26}  & \textbf{4.05}  & \textbf{0.95} & \textbf{0.89} & \textbf{0.85} & \underline{0.80} \\
\multirow{2}{*}{FedTSE (Q=1)} & RMSE    & \textbf{9.56}  & \textbf{8.02}  & \textbf{7.23}  & \underline{6.02}  & \underline{1.13} & \underline{1.01} & \underline{1.02} & \textbf{1.02} \\
                        & MAE     & \textbf{6.21}  & \textbf{5.30}  & \textbf{4.53}  & \underline{3.96}  & \underline{0.89} & \underline{0.80} & \underline{0.81} & \textbf{0.80}\\
		\hline
	\end{tabular}
\end{table}

\begin{figure}[!htbp]
     \centering
     \begin{subfigure}[b]{0.41\textwidth}
         \centering
         \includegraphics[width=\textwidth]{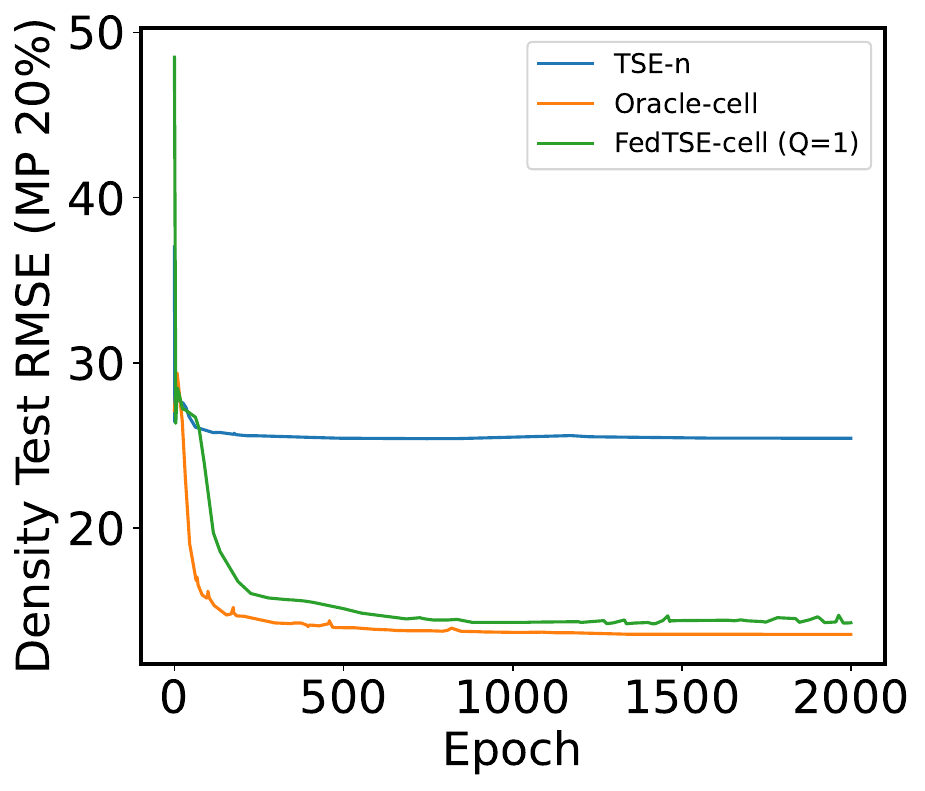}
         \caption{Density test RMSE (MP 20\% penetration rate)}
         \label{fig:loss1}
     \end{subfigure}
     \begin{subfigure}[b]{0.41\textwidth}
         \centering
         \includegraphics[width=\textwidth]{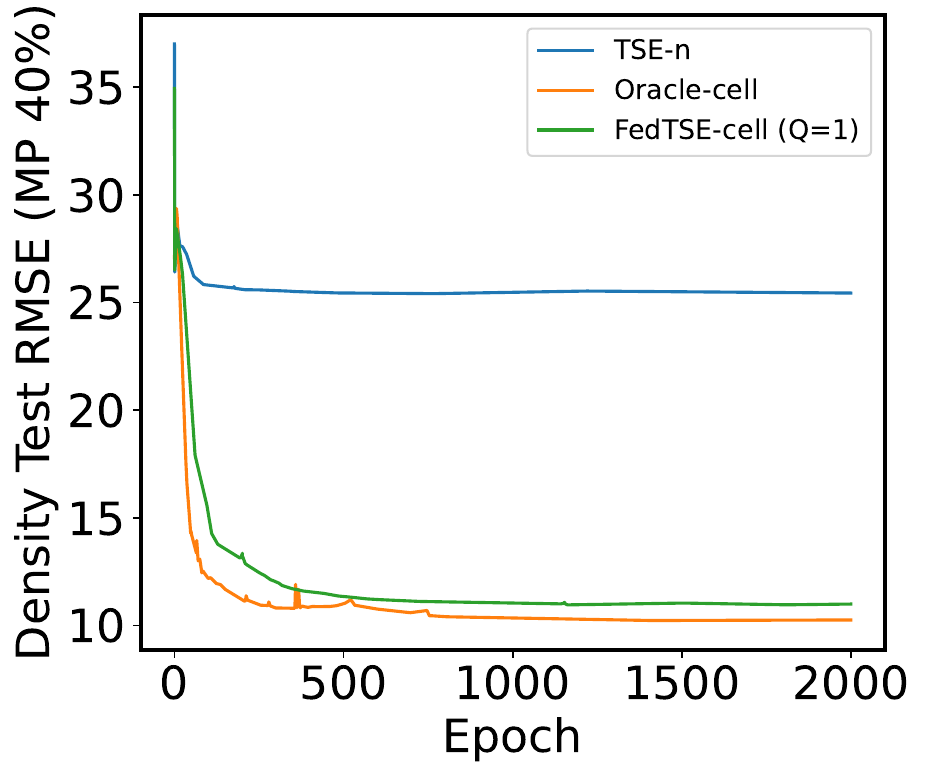}
         \caption{Density test RMSE (MP 40\% penetration rate)}
         \label{fig:loss2}
     \end{subfigure}
     \caption{Test RMSE curve for FedTSE}
    \label{fig:loss}
\end{figure}

\emph{Performance on simulation data}: Table \ref{sumotab1} summarizes the model performance on SUMO simulation data to perform more comprehensive evaluations. Specifically, two more benchmarks, Oracle-cell and FedTSE-cell, are added to evaluate the value of data fusion since the amount of real-world data is not sufficient for training these benchmarks. 
We have two new observations in the simulation data. 
First, we can see that, FedTSE-cell  yields a similar accuracy compared to Oracle-cell, which shows that the proposed vertical FL framework can lead to satisfactory tradeoff between privacy and estimation performance. Moreover, Figure~\ref{fig:loss} illustrates the test RMSE of density and flow during the training process, whereby the model performance is evaluated at each epoch.  From Figure~\ref{fig:loss}, we can see that FedTSE-cell not only has a similar performance to Oracle-cell at the end of training, the convergence speed is also not significant higher, which suggests that FedTSE-cell protects privacy at a marginal cost in model performance and training speed. 
Second, Oracle-cell and FedTSE-cell yield a much lower error compared to other benchmarks without cell representations. This demonstrates that higher-resolution features provided by MP can significantly improve data fusion. This is important because MP can be more willing to contribute higher resolution data if privacy is under protection, which shows the value of privacy protection in data fusion. 

\begin{table}[!htbp]
	\centering
	\caption{Model performance comparison on simulation dataset.}
	\label{sumotab1}
		\begin{tabular}{lccccccccc}
        \hline
        & Metrics & \multicolumn{4}{c}{Density (veh/km/lane)}         & \multicolumn{4}{c}{Flow (veh/min/lane)}     \\
                             \hline
MP Penetration rate (\%)     &         & 20    & 40    & 60    & 80    & 20   & 40   & 60   & 80   \\
\hline
\multirow{2}{*}{TSE-n}          & RMSE    & 14.19 & 14.19 & 14.19 & 14.19 & 0.96 & 0.96 & 0.96 & 0.96 \\
                             & MAE     & 6.47  & 6.47  & 6.47  & 6.47  & 0.75 & 0.75 & 0.75 & 0.75 \\
\multirow{2}{*}{TSE-p}          & RMSE    & 7.91  & 7.04  & 6.34  & 7.15  & 0.98 & 0.95 & 0.92 & 0.89 \\
                             & MAE     & 4.34  & 3.92  & 3.63  & 3.73  & 0.77 & 0.75 & 0.72 & 0.70 \\
\multirow{2}{*}{Oracle}          & RMSE    & 6.75  & 5.22  & 4.46  & 3.99  & 0.95 & 0.88 & 0.83 & 0.79 \\
                             & MAE     & 3.84  & 3.33  & 2.85  & 2.57  & 0.75 & 0.69 & 0.65 & 0.61 \\
\multirow{2}{*}{FedTSE (Q=1)}      & RMSE    & 6.81  & 5.70  & 5.04  & 4.27  & 0.92 & 0.90 & 0.86 & 0.87 \\
                             & MAE     & 3.94  & 3.57  & 3.19  & 2.90  & 0.72 & 0.71 & 0.67 & 0.68 \\
                             \hline
\multirow{2}{*}{Oracle-cell}     & RMSE    & \underline{5.65}  & \underline{3.84}  & \underline{3.20}  & \underline{2.65}  & \textbf{0.95} & \underline{0.85} & \underline{0.78} & \underline{0.69} \\
                             & MAE     & \underline{3.47}  & \underline{2.65}  & \underline{2.17}  & \underline{1.82}  & \textbf{0.75} & \underline{0.67} & \underline{0.61} & \underline{0.53} \\
\multirow{2}{*}{FedTSE-cell (Q=1)} & RMSE    & \textbf{5.76}  & \textbf{4.34}  & \textbf{4.02}  & \textbf{3.58}  & \underline{0.91} & \textbf{0.90} & \textbf{0.90} & \textbf{0.82} \\
                             & MAE     & \textbf{3.58}  & \textbf{3.02}  & \textbf{2.81}  & \textbf{2.48}  & \underline{0.71} & \textbf{0.71} & \textbf{0.70} & \textbf{0.64}\\
		\hline
	\end{tabular}
\end{table}

\subsubsection{Sensitivity analysis on the number of local updates $Q$}
\begin{figure}[!htbp]
     \centering
     \begin{subfigure}[b]{0.417\textwidth}
         \centering
         \includegraphics[width=\textwidth]{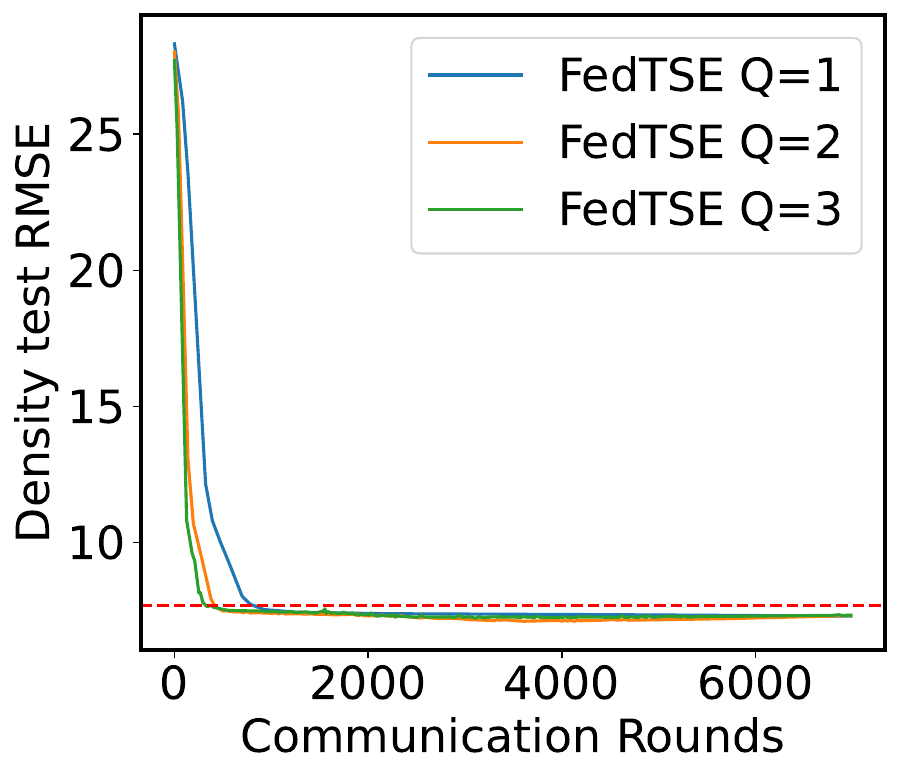}
         \caption{Density test RMSE curve (Red line equals to 7.7)}
         \label{fig:Q1}
     \end{subfigure}
     \begin{subfigure}[b]{0.401\textwidth}
         \centering
         \includegraphics[width=\textwidth]{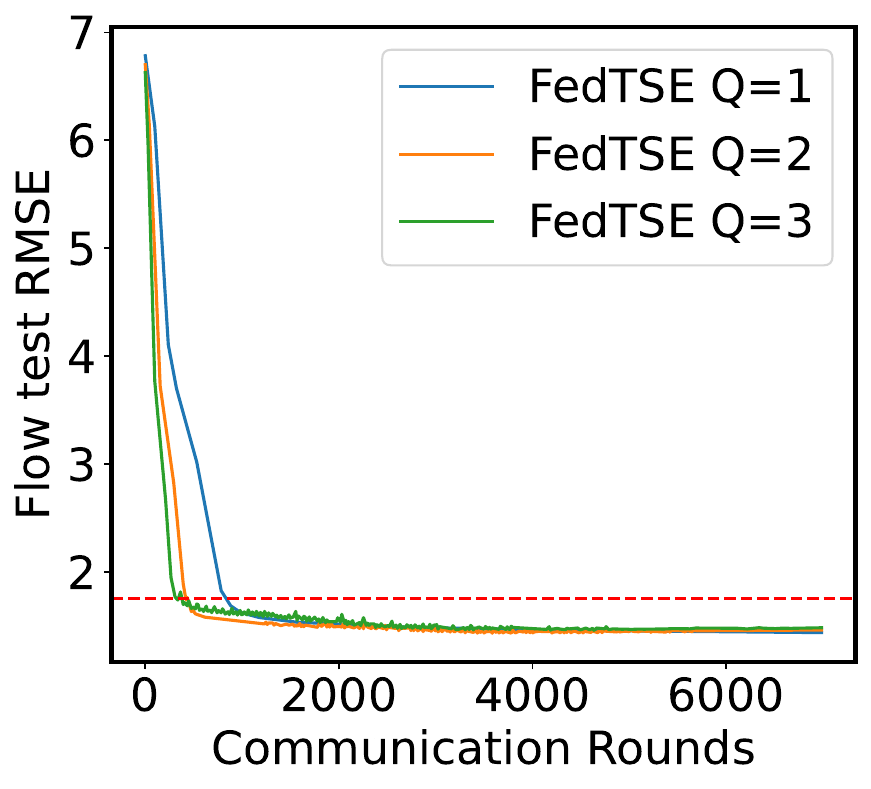}
         \caption{Flow test RMSE curve (Red line equals to 1.75)}
         \label{fig:Q2}
     \end{subfigure}
        \caption{Test RMSE curve for FedTSE with different Q}
        \label{fig:Q}
\end{figure}

In FedTSE, the parameter $Q$, representing the number of local updates, plays an important role in balancing the communication costs and estimation performance. Its value can significantly influence the speed of convergence and the overall effectiveness of the model. Table \ref{tab-Q} and Figure \ref{fig:Q} show that with a larger Q, the model converges faster and can achieve a certain loss threshold earlier. The test RMSE curve for flow is less stable when $Q = 3$, although reaches the threshold earlier. FedTSE (Q=2) is more stable and outperforms FedTSE (Q=1). Therefore, increasing the number of local updates appropriately can reduce the total number of communication rounds required to achieve better performance with FedTSE. This rapid convergence and improved model performance is critical, given that fewer communication rounds not only enhance the computational efficiency at the upper layer of the model but also minimize potential data leakage. 

\begin{table}[!htbp]
	\centering
	\caption{Number of communication rounds to reach a target RMSE. (One communication round corresponds to one mini-batch gradient descent of MA's global model with the MPs' intermediate output)}
	\label{tab-Q}
	\begin{tabular}{l c c}
		\hline
		 & Density test RMSE 7.7 & Flow test RMSE 1.75 \\
            & communication rounds   & communication rounds\\
		\hline
		FedTSE Q=1 & 812                   & 826    \\
        FedTSE Q=2 & 420                   & 420     \\
        FedTSE Q=3 & 308                    & 322     \\
		\hline
	\end{tabular}
\end{table}

\subsubsection{Sensitivity analysis on sample size}
This sensitivity analysis aims to leverage a well-calibrated SUMO simulation to inspect the impact of the sample size on the performance of FedTSE. The results are demonstrated in Table \ref{sampletab1}. It is worth noting that the simulation data yields similar performance to real-world data with the same sample size, which shows that the SUMO simulation is well-calibrated. Moreover, we can see that the RMSE of estimated density and flow decrease by 3.05 veh/km/lane and 0.28 veh/min/lane when the sample size increase from 600 to 3600. Hence, FedTSE is sensitive to the training sample size. 

\begin{table}[!htbp]
	\centering
	\caption{Sensitivity analysis on sample size.}
	\label{sampletab1}
        \small
		\begin{tabular}{lc|c|cccccc}
        \hline
                     &            & Real-world data & \multicolumn{6}{c}{Simulation data}     \\
                     \hline
\multicolumn{1}{c}{} & Sample size & 600             & 600  & 1200 & 1800 & 2400 & 3000 & 3600 \\
\hline
\multirow{2}{*}{Density (veh/km/lane)}   & RMSE       & 9.56            & 9.86 & 9.95 & 8.46 & 7.96 & 7.3  & 6.81 \\
                     & MAE        & 6.21            & 5.78 & 5.6  & 4.92 & 4.7  & 4.3  & 3.94 \\
\multirow{2}{*}{Flow (veh/min/lane)}   & RMSE       & 1.13            & 1.2  & 1.15 & 1.05 & 1.04 & 0.96 & 0.92 \\
                     & MAE        & 0.89            & 0.94 & 0.91 & 0.82 & 0.81 & 0.75 & 0.72\\
		\hline
	\end{tabular}
\end{table}

It is important to note that although there are many pilot studies that leverage drones to monitor traffic \citep{barmpounakis2020new}, the ground-truth labels are still challenging to obtain, and the real-world data suffers from limited sample size. Data collection via drone surveillance is expensive due to the limited battery capacity and detection range, as well as the public acceptance regarding the privacy risks of monitoring urban residents. Therefore, we cannot easily assume sufficient ground-truth data for training FedTSE. This motivates us to extend FedTSE to consider scenarios where real-world datasets are insufficient or even unavailable. To this end, we propose FedTSE-PI as presented in the following sections.

\section{FedTSE-PI: Physics-informed Vertical Federated Traffic State Estimation} \label{sec-fedtse-pi}
In this section, we extend the FedTSE proposed in Section \ref{sec-fedtse} to apply to scenarios where ground-truth labels are not available. This is a realistic consideration, as the collection of ground-truth labels can be expensive.
To address these issues, we introduce a Physics-Informed Vertical Federated Learning approach for TSE, hereafter named FedTSE-PI, that integrates models in traffic flow theory, e.g., the Cell Transmission Model, with FedTSE to enhance data efficiency while preserving traffic data privacy. 
Such an approach combines the data efficiency of traffic flow models and the privacy-preserving capabilities of FedTSE. Specifically, instead of requiring ground-truth labels, this approach adopts a loss function that combines traffic model information with only critical partial observations, such as traffic flow on road links equipped with loop detectors (provided by MA) and link-level speed information of MPs' fleets (provided by MPs). To prevent MPs sensitive information from being inferred from the sharing of these partial observations, we further propose a privacy-preserving mechanism for MA to calculate the gradients of the loss function without requiring MPs to explicitly share their observations. 

This section is organized as follows. Section~\ref{sec 4-1} presents the general framework of FedTSE-PI and Section~\ref{sec 4-2} presents the privacy-preserving training algorithm. 

\subsection{General framework of FedTSE-PI} \label{sec 4-1}
To enhance the data efficiency of training FedTSE, we leverage a promising framework of physics-informed neural networks (PINNs) \citep{shi2021physics, di2023physics} to integrate traffic flow theory (e.g., kinematic wave theory) into FedTSE. 
The advantage of traffic flow models is that they have a clear physical meaning and can provide relatively accurate estimates of traffic states with only a small number of parameters to calibrate (e.g., parameters for the fundamental diagrams). Extensive research has proposed TSE approaches based on traffic flow models using statistical filters or optimization-based methods~\citep{seo2017traffic,makridis2023adaptive,nie2023correlating,lu2023physics}, which can leverage critical partial observations, such as traffic flow on road links equipped with loop detectors (provided by MA) and speed information of MP's fleet (provided by MP). 
However, these traffic flow-based approaches  assume that MA has access to all these partial observations without considering the privacy concerns of MPs. 
Existing privacy-preserving physics-based TSE methods generally are generally based on open-loop data perturbation~\citep{le2014real,he2020optimal}, which can hinder data quality and hence TSE performance. 
In contrast, FedTSE can effectively protect the data privacy of both MA and MPs while ensuring each party uses the true data. We aim to combine traffic models with FedTSE to exploit the benefits of both approaches. 

\begin{figure}[!htbp]
	\centering
	\includegraphics[width=0.7\textwidth]{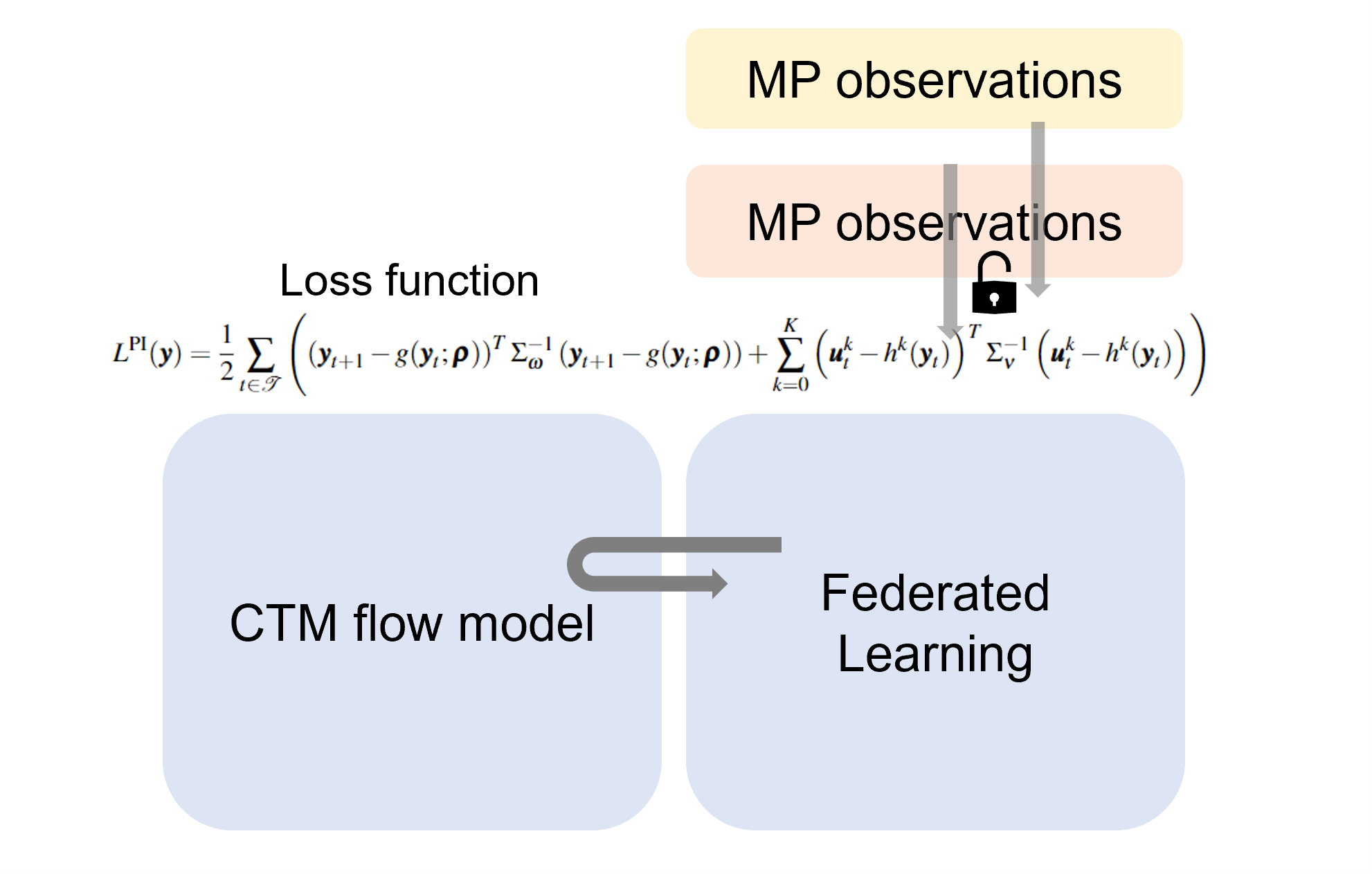}
	\caption{Schematic illustration of FedTSE-PI}
	\label{FedTSE-PI}
\end{figure}

Figure~\ref{FedTSE-PI} shows the general framework, where the traffic models are used to specifically design a loss function that can allow MA and MPs to train  the parameters of sub-models $\bm{\phi}^k$ using only critical partial observations instead of ground-truth labels.  

We assume that MA possesses a traffic model that can be mathematically described by a state-space model relying on the following recursive equation.  
 \begin{align} \label{eq:ctm_ss}
    \bm{y}_{t+1} = \bm{g}(\bm{y}_t; \bm{\rho}) + \bm{\omega}_t 
\end{align}
where $\bm{y}_t$ represents the state including flow and density\footnote{Notice that we use symbol $y$ instead of the commonly used $x$ to be consistent with our previous definitions of traffic states in Section \ref{sec-fedtse}.},   $\bm{\rho}$ represents model parameters that can include parameters in the fundamental diagram, signal timings, and estimated turning ratios, and $\bm{\omega}_t$ represents the model noises. 
Without loss of generality, we assume that these parameters and model noise distribution are calibrated \emph{a priori}, which can be realistic for cities with a well-calibrated simulator for their transportation systems. If the parameters and model noise distribution are unknown, we can also estimate them simultaneously with the training of FL models. 

The critical partial observations of both MA and MPs can be seen as measurements of the model, represented by $\bm{u}_t^k$: 
\begin{align} \label{eq:ctm_measure}
    \bm{u}_t^k = \bm{h}^k(\bm{y}_t) + \bm{\nu}_t^k
\end{align}
where $\bm{h}^k(\cdot)$ represents the measurement function, and $\bm{u}_t^k$ represents the measurement noises of data owner $k$. For MA, the partial observations can be flow measured by loop detectors, and hence $\bm{h}^k(\cdot)$ can be a linear function with a coefficient matrix consisting of only 1 and 0, whereby an element is 1 if the corresponding state is flow and the corresponding link has loop detectors installed. For MPs, these partial observations can be the link-level speed information provided by their vehicle trajectories, for which $\bm{h}^k(\cdot)$ can be defined as the ratio between flow and density. 
We assume that the form of the measurement function $\bm{h}^k(\cdot)$ and the distribution of noises $\bm{\nu}_t^k$ are publicly known by both MPs and MA, since it is natural to specify this information as a key requirement before collaboration.  
    
Therefore, given measurements $\{\bm{u}_t^k\}_{t\in\mathcal{T},k\in\{0,\cdots,K\}}$, MA is interested in solving $\bm{y} = \{\bm{y}_t\}_{t\in\mathcal{T}}$ maximizing the likelihood represented as  
\begin{align} \label{eq:likelihood}
\max_{\bm{y}}~ P\left(\{\bm{y}_t, \bm{u}_{t}\}_{t\in\mathcal{T}}\right) \Leftrightarrow  \max_{\bm{y}} ~ \prod_{t\in\mathcal{T}}P(\bm{y}_{t+1}|\bm{y}_t) P(\bm{u}_{t}|\bm{y}_t) 
\end{align}

With the assumption that both the modeling and measurement noises follow Gaussian distributions, i.e.,  $\bm{\omega}_t\overset{\mathrm{iid}}{\sim} N(\bm{0}, \Sigma_{\omega})$ and $\bm{\nu}_t^k\overset{\mathrm{iid}}{\sim} N(\bm{0}, \Sigma_{\nu})$, maximizing the likelihood is equivalent to minimizing the following loss function 
\begin{align}
    L^{\rm{PI}}(\bm{y}) = \frac{1}{2}\sum_{t\in\mathcal{T}}\left(
    \left(\bm{y}_{t+1} - \bm{g}(\bm{y}_t; \bm{\rho})\right)^T\Sigma_{\omega}^{-1}\left(\bm{y}_{t+1} - \bm{g}(\bm{y}_t; \bm{\rho})\right) + 
    \sum_{k=0}^K\left( \bm{u}_t^k - \bm{h}^k(\bm{y}_t) \right)^T\Sigma_{\nu}^{-1}\left( \bm{u}_t^k - \bm{h}^k(\bm{y}_t) \right) 
    \right) \label{eq:ctm_loss}
\end{align}

Notice that in the FedTSE framework, $\bm{y}=\bm{y}(\bm{\theta})$ is computed by a combination of sub-models of all data owners, each parameterized by parameters $\bm{\theta} = \{\bm{\theta}^k\}_{k=0}^K$. Hence, the loss function $L^{\rm{PI}}(\bm{y}(\bm{\theta}))$ can be used as the loss function for training these sub-models. 

We specifically employ the Cell Transmission Model (CTM), a numerical method that discretizes time and space to solve LWR equations \citep{daganzo1994cell},  to characterize the dynamics of traffic flow. 
Note that we use CTM as an example due to its popularity and simplicity, and our methodological framework can accommodate any model with a recursive structure, such as the store-and-forward model \citep{aboudolas2009store}, METANET \citep{papageorgiou1990modelling}, and Macroscopic Fundamental Diagram (MFD)-based dynamic models for network-level traffic state estimation~\citep{geroliminis2008existence}. 

The urban transportation network is discretized into a set of cells with length $\Delta x$ such that $\Delta x > v^f\Delta t$, where $v^f$ indicates the free-flow speed of the link. For presentation brevity, we assume each cell corresponds to a road link $n\in\mathcal{V}$, as we can always divide a link into multiple cells. 

Then the flow conservation can be represented by Equation \ref{421}. 
\begin{align}\label{421}
    k_{n,t+1} = k_{n,t}+ \frac{\Delta t}{\Delta x}\left(\sum_{m: (m,n)\in \mathcal{E}} q_{mn,t} - \sum_{m: (n,m)\in \mathcal{E}} q_{nm,t}\right)
\end{align}
where $q_{mn,t}$ represents the transfer flow from cell $m$ to its adjacent cell $n$. 

Given the turning ratio $p_{mn}$ along the edge $(m,n) \in \mathcal{E}$, the sending function describing the maximum flow that can leave cell $m$ for cell $n$ can be written as
\begin{align}\label{422}
    S_{mn,t} = p_{mn}\min \{q_m^c, v_m^f k_{m,t}\}
\end{align}
where $q_m^c$ is the capacity of a cell, and $v_m^f$ is the free-flow speed. The sending function is calculated as the minimum of the capacity and the flow that desires to leave the cell.   

The receiving function defines the maximum flow that can be accommodated by cell $n$, which can be written as
\begin{align}\label{424}
    R_{n,t}=\min \{w_n(k_n^{\text{jam}}-k_{n,t}),q_n^c\}
\end{align}
where $k_n^{\text{jam}}$ represents the jam density and $w_n$ the backward wave speed for cell $n$. 

The flow that can enter cell $n$ can be calculated as the minimum of total sending flow and receiving flow 
\begin{align}\label{425}
    Q_{n,t} = \min\left\{\sum_{m:(m,n)\in\mathcal{E}}S_{mn,t}, R_{n,t}\right\}
\end{align}

Then, assuming the traffic within each cell is homogeneous, the transfer flow between cell $m$ to cell $n$ can be derived as Equation \ref{426}: 
\begin{align}\label{426}
    q_{mn,t} = Q_{n,t} \frac{S_{mn,t}}{\sum_{m':(m',n)\in\mathcal{E}S_{m'n,t}}}
\end{align}

We can summarize CTM as a recursive model in the form of Equation~\ref{eq:ctm_ss}, where the state variables $\bm{y}_t = (\{k_{nt}\}_{n\in\mathcal{V}},\{q_{mnt}\}_{(m,n)\in\mathcal{E}})$ is the state,  and the model parameters can be represented by the parameters of the fundamental diagrams of all links, i.e., $\bm{\rho}=\left\{v_n^f,k_n^{\text{jam}},q_n^c\right\}_{n\in\mathcal{V}}$. We incorporate CTM into the loss function Equation \ref{eq:ctm_loss} to facilitate the training of FedTSE.

\begin{algorithm}[hbt!]
\caption{Algorithm for FedTSE-PI}\label{alg:2}
\DontPrintSemicolon
\SetAlgoLined
\textbf{Input:} Private datasets $\mathcal{D}^k$ and learning rate $\eta_k$ for data owner $k = 0, 1, \ldots, K$, regularizer weight parameter $\lambda$, and the number of local updates $Q$\\
\textbf{Output:} $\Theta = (\bm{\theta}^0, \bm{\theta}^1, \ldots, \bm{\theta}^k)$\\

\For{each iteration $i = 1, 2, \ldots$}{
    \If{ $i\mod Q = 0$}{
        MA randomly samples a mini-batch of time steps $\mathcal{B} \in \mathcal{T}$ and synchronizes it with MPs\;
        \For{$k = 1, \ldots, K$ in parallel}{
            MP $k$ computes local output $\bm{z}^k_t = \bm{\phi}^k(\bm{x}^k_t; \bm{\theta}^k)$ with its private sub model for $t \in \mathcal{B}$\;
            MP $k$ sends ${\bm{z}^k_t}$ to MA\;
        }

        \For{$k = 1, \ldots, K$ in parallel}{
            MA and MP $k$ jointly computes $\Delta_{\bm{\theta}}^k =  \sum_{t \in \mathcal{B}}(\bm{u}_t^k)^T\Sigma_{\nu}^{-1} \frac{\partial \bm{h}^k}{\partial \bm{y}}\frac{\partial \bm{y}}{\partial \bm{\theta}^0}$ and $\Delta_{\bm{z}}^k =  \sum_{t \in \mathcal{B}}(\bm{u}_t^k)^T\Sigma_{\nu}^{-1} \frac{\partial \bm{h}^k}{\partial \bm{y}}\frac{\partial \bm{y}}{\partial \bm{\bm{z}^k_t}}$ using secure functional encryption-based inner production computation \;
        }
        MA uses $\{\Delta_{\bm{z}}^k\}_{k=0}^K$ and $\{\Delta_{\bm{\theta}}^k\}_{k=0}^K$ to calculate gradients $\frac{\partial L}{\partial \bm{\theta}^0}$ and $\frac{\partial L}{\partial \bm{z}^k_t}$ according to Equation \ref{eq:pi_gradients}\;          
        MA updates sub-model parameters $\theta_0^{i+1}=\theta_0^i-\eta_0 \frac{\partial L}{\partial \bm{\theta}^0} $ and sends $\frac{\partial L}{\partial \bm{z}^k_t}$ to each MP\; 
    }

    \For{$k = 1, \ldots, K$ in parallel}{
        MP $k$ computes $\frac{\partial L}{\partial \bm{\theta}^k}$ with the most recent $\frac{\partial L}{\partial \bm{z}^k_t}$ according to Equation \ref{gradient}\;
        MP $k$ Update $\theta_k^{i+1}=\theta_k^i-\eta_k \frac{\partial L}{\partial \bm{\theta}^k}$\;
    }
    \If{convergence criterion met}{
        break\;
    }
    }
\end{algorithm}

\subsection{Privacy-preserving training of FedTSE-PI}\label{sec 4-2}

The training process of FedTSE-PI is summarized in Algorithm~\ref{alg:2}, which follows the training of FedTSE. MA calculates two types of gradients: (1) the gradient of $L^{\rm{PI}}$ with respect to the model parameters of MA ($\bm{\theta}^0$), i.e., $\frac{\partial L^{\rm{PI}}}{\partial \bm{\theta}^0}$, which will be used to update its parameters $\bm{\theta}^0$, and 
(2) the gradients of $L^{\rm{PI}}$ with respect to the output $z_t^k$, i.e., $\frac{\partial L^{\rm{PI}}}{\partial \bm{z}^k_t}$, which will be sent back to MP $k$ for MP $k$ to update its parameters. 

However, one key difference between the training of FedTSE-PI and FedTSE lies in the loss function, whereby the loss function of FedTSE relies on the ground-truth labels collected by MA, whereas the loss function of FedTSE-PI requires critical partial observations made by both MA and MPs. In other words, the key input to the loss function of FedTSE-PI is distributed among MA and MPs, and MPs may be reluctant to explicitly share such information due to privacy considerations. 

To address such a privacy issue, we propose a privacy-preserving gradient calculation mechanism that allows MA to calculate the gradients without MPs having to explicitly share their data. This mechanism relies on secure functional encryption~\citep{lewko2010fully} that encrypts private data into an encrypted message, such that any user with the encrypted message can learn a predetermined function about the private data but nothing else. We specifically leverage inner product encryption (IPE) that allows the computation of the inner products of two vectors $\bm{a}_1$ and $\bm{a}_2$ without sharing any other information other than their inner product $\bm{a}_1^T\bm{a}_2$. Inner production encryption has been extensively investigated in cryptography literature~\citep{lewko2010fully,okamoto2015achieving,couteau2022non}, which ensures the fast and secure computation of inner products of two vectors, which can be possessed by two parties. Our mechanism can employ any existing two-party inner production encryption approaches. Instead of introducing the details of these approaches, we focus on how to formulate the gradient calculation problem into a series of secure inner production calculation problems. 

Let us consider a minibatch $\mathcal{B}\subset\mathcal{T}$. We take the example of calculating $\frac{\partial L^{\rm{PI}}}{\partial \bm{\theta}^0}$, and the calculation of the $\frac{\partial L^{\rm{PI}}}{\partial \bm{z}^k_t}$ is similar. By the chain rule of partial derivatives, we can obtain
\begin{align}
    \frac{\partial L^{\rm{PI}}}{\partial \bm{\theta}^0} & =  
    \sum_{t \in \mathcal{B}}\frac{\partial L^{\rm{PI}}}{\partial \bm{y}_t}
    \frac{\partial \bm{y}_t}{\partial \bm{\theta}^0} \notag \\
    = & \frac{1}{2}\sum_{t \in \mathcal{B}} \frac{\partial}{\partial \bm{y}_t }
    (\bm{y}_{t+1} - \bm{g}(\bm{y}_t; \bm{\rho}))^T\Sigma_{\omega}^{-1}(\bm{y}_{t+1} - \bm{g}(\bm{y}_t; \bm{\rho})) 
    \frac{\partial \bm{y}_t}{\partial \bm{\theta}_0} \notag \\
    & + \frac{1}{2}\sum_{t \in \mathcal{B}}\sum_{k=0}^K\frac{\partial}{\partial \bm{y}_t} ( \bm{u}_t^k - \bm{h}^k(\bm{y}_t))^T\Sigma_{\nu}^{-1}(\bm{u}_t^k - \bm{h}^k(\bm{y}_t)) \frac{\partial \bm{y}_t}{\partial \bm{\theta}_0} \notag \\
    = & \frac{1}{2}\sum_{t \in \mathcal{B}} \frac{\partial}{\partial \bm{y}_t }
    (\bm{y}_{t+1} - \bm{g}(\bm{y}_t; \bm{\rho}))^T\Sigma_{\omega}^{-1}(\bm{y}_{t+1} - \bm{g}(\bm{y}_t; \bm{\rho})) 
    \frac{\partial \bm{y}_t}{\partial \bm{\theta}_0} \notag \\
    & - \sum_{t \in \mathcal{B}}\sum_{k=0}^K
    ( \bm{u}_t^k - \bm{h}^k(\bm{y}_t))^T \Sigma_{\nu}^{-1} \frac{\partial \bm{h}^k }{\partial \bm{y}_t} 
    \frac{\partial \bm{y}_t}{\partial \bm{\theta}_0} \notag \\
    = & \frac{1}{2}\sum_{t \in \mathcal{B}} \frac{\partial}{\partial \bm{y}_t }
    (\bm{y}_{t+1} - \bm{g}(\bm{y}_t; \bm{\rho}))^T\Sigma_{\omega}^{-1}(\bm{y}_{t+1} - \bm{g}(\bm{y}_t; \bm{\rho})) 
    \frac{\partial \bm{y}_t}{\partial \bm{\theta}_0} \notag \\
    & - \sum_{t \in \mathcal{B}}\sum_{k=0}^K
   (\bm{u}_t^k)^T \Sigma_{\nu}^{-1} \frac{\partial \bm{h}^k }{\partial \bm{y}_t} 
    \frac{\partial \bm{y}_t}{\partial \bm{\theta}_0} 
    + \sum_{t \in \mathcal{B}}\sum_{k=0}^K (\bm{h}^k(\bm{y}_t))^T \Sigma_{\nu}^{-1} \frac{\partial \bm{h}^k }{\partial \bm{y}_t} 
    \frac{\partial \bm{y}_t}{\partial \bm{\theta}_0} 
    \label{eq:pi_gradients}
 \end{align}
where the first and third terms are known to MA, but the second term includes the private data of each MP $k$, i.e., $\bm{u}_t^k$. Nevertheless, we notice that the second term can be represented as a combination of a sequence of inner products between $\bm{u}_t^k$ and each column vector in matrix $\Sigma_{\nu}^{-1} \frac{\partial \bm{h}^k}{\partial \bm{y}}
    \frac{\partial \bm{y}}{\partial \bm{\theta}^0}$. 
    Hence, we can use inner production encryption to allow secure calculation of the second term without having to enclose information from both parties. 

By integrating with inner product encryption, the gradients can be calculated without MPs having to share their data. 
Nevertheless, it is still possible for adversaries to infer the true values of $\bm{\mu}_t^k$ by solving a linear equation if they know the inner product between $\bm{\mu}_t^k$ and a sequence of vectors. This can be prevented with several tricks. First, the mini-batch size can be chosen as sufficiently large such that the dimension of $\{\bm{u}_t^k\}_{t \in \mathcal{B}}$ is higher than the rank of matrix $\Sigma_{\nu}^{-1} \frac{\partial \bm{h}^k}{\partial \bm{y}}
    \frac{\partial \bm{y}}{\partial \bm{\theta}^0}$, which ensures that the values of $\bm{\mu}_t^k$ cannot be inferred from inner products. This can be, in reality, practical since the availability of partial observations is abundant (in contrast to the limited availability of ground-truth data), with the continuous measurements of MA's detectors and operations of MPs' fleets.  
Second, MA can collaborate with multiple MPs such that each MP can report critical partial observations on a subset of links traversed by its fleet, which helps protect MPs' data privacy by not using the data.

\section{Case Study on FedTSE-PI} \label{sec-fedtse-pi-case}
\subsection{Benchmarks for FedTSE-PI}
To evaluate the performance of FedTSE-PI, we compare the following benchmarks, where MA is assumed to have no access to ground-truth traffic states, and hence data fusion refers to the fusion of both features and measurements. 

\begin{itemize}

    \item \textbf{QEST (physics-based density estimation with MP's data)}: This benchmark 
    is a state-of-the-art physics-based TSE approach proposed by \citet{yang2018queue}, which estimates queue profile from trajectory data provided by MP by solving a convex optimization problem formulated using kinematic wave theory. MP further uses the queue profile to produce its density measurements. Note that this benchmark does not use MA's data. It is worth noting that the embedded convex optimization is efficient to solve, costing no more than 2s on a PC with a AMD Ryzen 5 3600XT 6-Core Processor for each signal phase. Hence, embedding QEST with the training of learning-based TSE methods (e.g., FedTSE-PI) does not cost significant computation complexity. 
    
    \item \textbf{QEST-f (physics-based density estimation with data fusion)}: This benchmark is an extension to QEST to include MA's loop detector and signal timing data, in which MA sends these data to MP, and MP produces density measurements by solving a convex optimization problem. Note that this benchmark could undermine the privacy of MA. This benchmark is used to evaluate the advantage of learning-based approaches in comparison to physics-based approaches. 
    
    \item \textbf{TSE-PI-p (physics-informed partial data fusion without privacy protection)}: 
    This benchmark is the physics-informed version of TSE-p, where MA trains physics-informed neural networks with MP's speed information. The features of MA and MP are identical to those for TSE-p. The loss function follows Eq.(\ref{eq:ctm_loss}) with the measurements of MP being the speed information. Note that similar to TSE-p, privacy is not explicitly protected for this benchmark, and MP only wants to share speed information due to privacy concerns. 

    \item \textbf{Oracle-PI (physics-informed data fusion without privacy protection)}: This model is the physics-informed version of Oracle, whereby MP fully trusts MA, and MA trains physics-informed neural networks with its own data and MP's data without privacy protection.  The features of MA and MP are identical to those for Oracle. The loss function follows Eq.(\ref{eq:ctm_loss}) with the measurements of MP being the estimated speed and density using QEST. 

    \item \textbf{UKF (Unscented Kalman Filter)}: This benchmark is another state-of-the-art physics-based approach proposed by \citep{makridis2023adaptive} that fuses trajectory data with loop detector data without considering privacy concerns. 

    \item \textbf{FedTSE-PI}: the proposed physics-informed privacy-preserving data fusion vertical FL framework. The features of MA and MP are identical to those for FedTSE. The loss function follows Eq.(\ref{eq:ctm_loss}) with the measurements of MP being the speed information and estimated density using QEST. Note that here, the privacy is protected in two perspectives. First, FL ensures that data is distributed. Second, encryption-based inner product computation ensures that MP's speed and density measurements are incorporated into the loss function with privacy protection. 
    
    \item \textbf{FedTSE-PI-v}: This benchmark is identical to FedTSE-PI, except the loss function follows Eq.(\ref{eq:ctm_loss}) with the measurements of MP being only the speed information. 

    \item \textbf{FedTSE-PI-k}: This benchmark is identical to FedTSE-PI, except the loss function follows Eq.(\ref{eq:ctm_loss}) with the measurements of MP being only the estimated density using QEST. 
    
\end{itemize}

Note that in practice, although ground-truth labels are limited, the measurements of MA and MPs (e.g., loop detector data and trajectory data) are abundant, as these data owners continuously generate operational data. Therefore, it is reasonable to assume that both MA and MPs have a long history of loop detector data and trajectory data. To realistically replicate this, we use simulation data to evaluate the performance of these benchmarks due to the limited data provision of real-world data.  The experiment settings are the same with Section \ref{sec:setting_fedtse}.

\subsection{Performance of FedTSE-PI}
\begin{table}[!htbp]
	\centering
	\caption{Model performance comparison of FedTSE-PI. (\textbf{F} represents whether there exists data fusion. \textbf{P} represents whether the data fusion is privacy-preserving. \textbf{V} and \textbf{K} represent whether MP shares speed or density measurements respectively.)}
	\label{tab-pi}
        \small
		\begin{tabular}{c|c|c|c|lccccccccc}
        \hline
        \multicolumn{4}{l}{Key factors} && Metrics &\multicolumn{4}{c}{Density (veh/km/lane)} &\multicolumn{4}{c}{Flow (veh/min/lane)} \\
        \hline
        \textbf{F} & \textbf{P}& \textbf{V}&\textbf{K}&Pr (\%)&&20 &40 &60 &80 &20 &40 &60 &80 \\
        \hline
        \multirow{2}{*}{} &\multirow{2}{*}{} &\multirow{2}{*}{\checkmark} &\multirow{2}{*}{\checkmark} &\multirow{2}{*}{QEST} &RMSE &8.20 &7.05 &6.64 &6.43 &5.61 &5.22 &5.06 &4.90 \\
        &&&&&MAE &6.20 &5.44 &5.15 &5.04 &3.52 &3.24 &3.14 &3.04 \\
        \hline
        \multirow{2}{*}{\checkmark} &\multirow{8}{*}{} &\multirow{2}{*}{\checkmark} &\multirow{2}{*}{\checkmark} &\multirow{2}{*}{QEST-f} &RMSE &6.89 &5.77 &5.38 &5.23 &4.39 &4.20 &4.12 &4.08 \\
        &&&&&MAE &4.94 &4.34 &4.11 &4.06 &2.47 &2.37 &2.34 &2.32 \\
        \multirow{2}{*}{\checkmark}&&\multirow{2}{*}{\checkmark} &\multirow{2}{*}{} &\multirow{2}{*}{TSE-PI-p} &RMSE &11.91 &11.30 &11.01 &11.01 &5.15 &4.86 &4.72 &4.64 \\
        &&&&&MAE &8.11 &7.46 &7.13 &6.96 &3.53 &3.24 &3.09 &2.86 \\
        \multirow{2}{*}{\checkmark}&&\multirow{2}{*}{\checkmark} &\multirow{2}{*}{\checkmark} &\multirow{2}{*}{Oracle-PI}& RMSE &\underline{6.32} &\underline{5.58} &\underline{4.93} &\underline{4.85} &\textbf{3.43} &\textbf{3.09} &\underline{2.77} &\underline{2.59} \\
        &&&&&MAE &\underline{4.64} &\underline{4.13} &\underline{3.71} &\underline{3.63} &\textbf{2.54} &\textbf{2.24} &\underline{1.94} &\underline{1.79} \\
        \multirow{2}{*}{\checkmark}&&\multirow{2}{*}{\checkmark} &\multirow{2}{*}{\checkmark} &\multirow{2}{*}{UKF} &RMSE &7.95 &6.88 &6.57 &6.29 &4.62 &4.38 &4.21 &4.14 \\
        &&&&&MAE &6.14 &5.41 &5.20 &5.02 &2.56 &2.47 &2.41 &2.36 \\
        \hline
        \multirow{2}{*}{\checkmark} & \multirow{2}{*}{\checkmark} & \multirow{2}{*}{\checkmark} & \multirow{2}{*}{} & \multirow{2}{*}{FedTSE-PI-v} & RMSE & 10.35 & 10.24 & 10.58 & 10.71 & 4.33 & 4.17 & 3.89 & 4.13 \\
        &&&&& MAE &6.80 &6.54 &6.45 &6.56 &2.85 &2.74 &2.31 &2.40 \\
        \multirow{2}{*}{\checkmark}&\multirow{2}{*}{\checkmark}&\multirow{2}{*}{} &\multirow{2}{*}{\checkmark} &\multirow{2}{*}{FedTSE-PI-k} &RMSE &7.25 &6.25 &6.21 &5.90 &3.62 &3.66 &3.48 &3.59 \\
        &&&&&MAE &5.40 &4.70 &4.67 &4.52 &2.60 &2.63 &2.54 &2.58 \\
        \multirow{2}{*}{\checkmark}&\multirow{2}{*}{\checkmark}&\multirow{2}{*}{\checkmark} &\multirow{2}{*}{\checkmark} &\multirow{2}{*}{FedTSE-PI} &RMSE &\textbf{6.37} &\textbf{5.71} &\textbf{5.07} &\textbf{5.05} &\underline{3.31} &\underline{3.04} &\textbf{2.82} &\textbf{2.83} \\
        &&&&&MAE &\textbf{4.81} &\textbf{4.23} &\textbf{3.81} &\textbf{3.72} &\underline{2.39} &\underline{2.18} &\textbf{1.98} &\textbf{2.02}\\
        \hline
\end{tabular}
\end{table}

\begin{figure}[!htbp]
     \centering
     \begin{subfigure}[b]{0.49\textwidth}
         \centering
         \includegraphics[width=\textwidth]{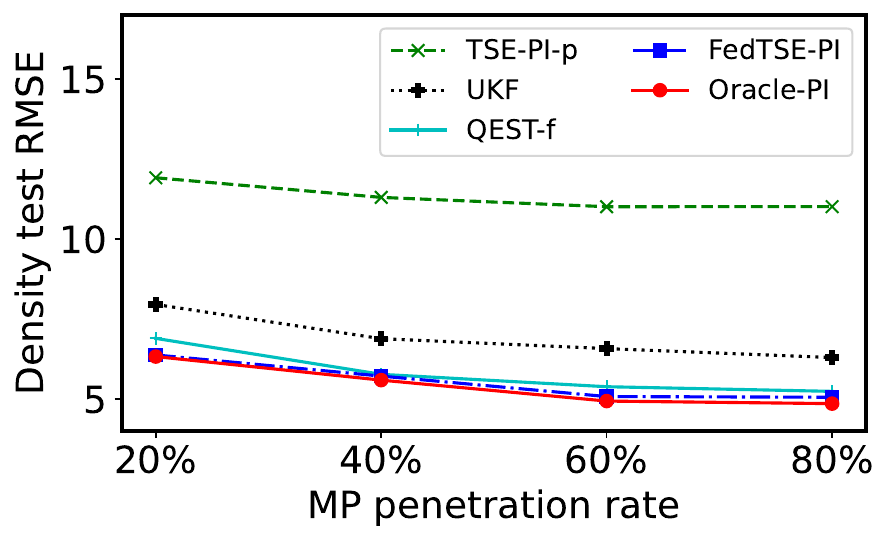}
         \caption{Density RMSE}
         \label{fig:pi-pr1}
     \end{subfigure}
     \hfill
     \begin{subfigure}[b]{0.48\textwidth}
         \centering
         \includegraphics[width=\textwidth]{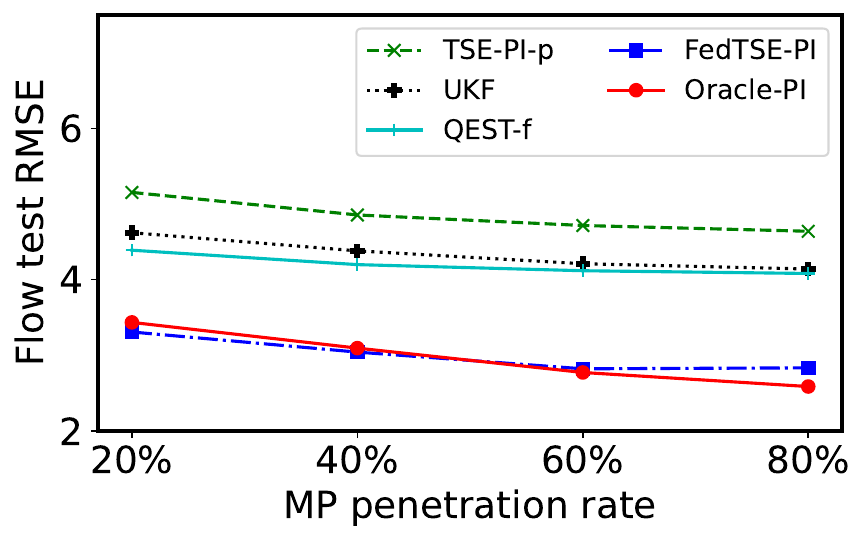}
         \caption{Flow RMSE}
         \label{fig:pi-pr2}
     \end{subfigure}
        \caption{Model performance comparison with different MP penetration rate.}
        \label{fig:pi-pr}
\end{figure}

Table \ref{tab-pi} and Figure \ref{fig:pi-pr} compare the performance of benchmarks using simulation data. The benchmarks with the best estimation results are \underline{underlined}, while those ranked second are highlighted in \textbf{bold}. 

\noindent \textbf{Tradeoff between privacy and estimation performance}. We evaluate the tradeoff between privacy and estimation performance of FedTSE-PI by comparing it to (i) state-of-the-art physics-based TSE algorithms with data fusion, including QEST-f and UKF, and (ii) Oracle-PI without considering privacy. We can see that even though a privacy-preserving mechanism is involved, FedTSE-PI still outperforms these physics-based methods and yield a similar estimation accuracy as Oracle-PI (i.e., with at most 4\% difference). This shows that FedTSE-PI can protect privacy at a marginal cost on estimation performance. 

The advantage of FedTSE-PI is further highlighted in Figure \ref{fig:pi-pr}, where the blue line (i.e., FedTSE-PI) outperforms all other existing methods for TSE and is close to the red line (Oracle-PI). This suggests that the integration of the physics model and FL could better capture the traffic dynamics, showing the benefits of incorporating learning-based components. 
Overall, despite having no access to ground-truth labels, FedTSE-PI can achieve excellent performance on the TSE problem.

\begin{figure}[!htbp]
     \centering
     \begin{subfigure}[h]{0.7\textwidth}
         \centering
         \includegraphics[width=\textwidth]{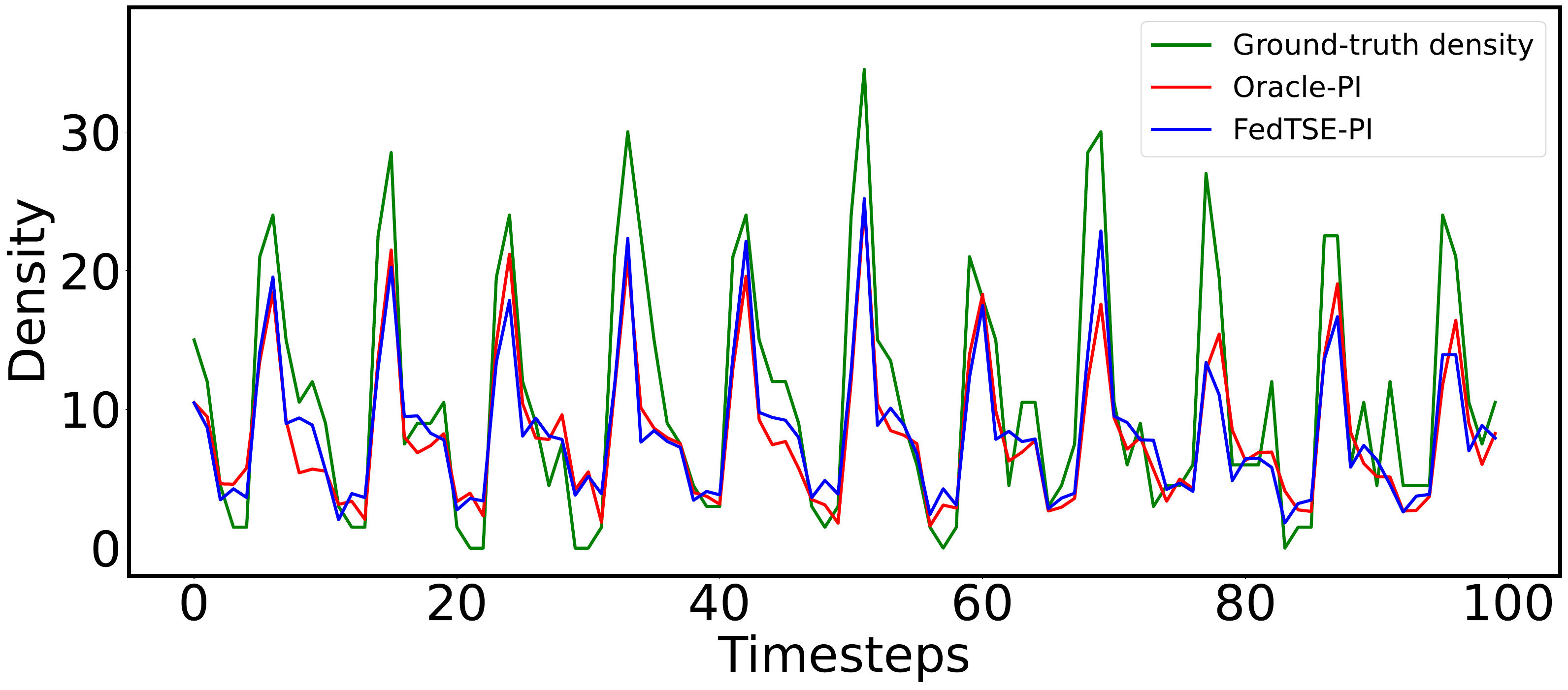}
         \caption{Density estimation results of one link with no loop detector (veh/km/lane).}
         \label{fig:pi-est-1}
     \end{subfigure}
     \hfill
     \begin{subfigure}[h]{0.7\textwidth}
         \centering
         \includegraphics[width=\textwidth]{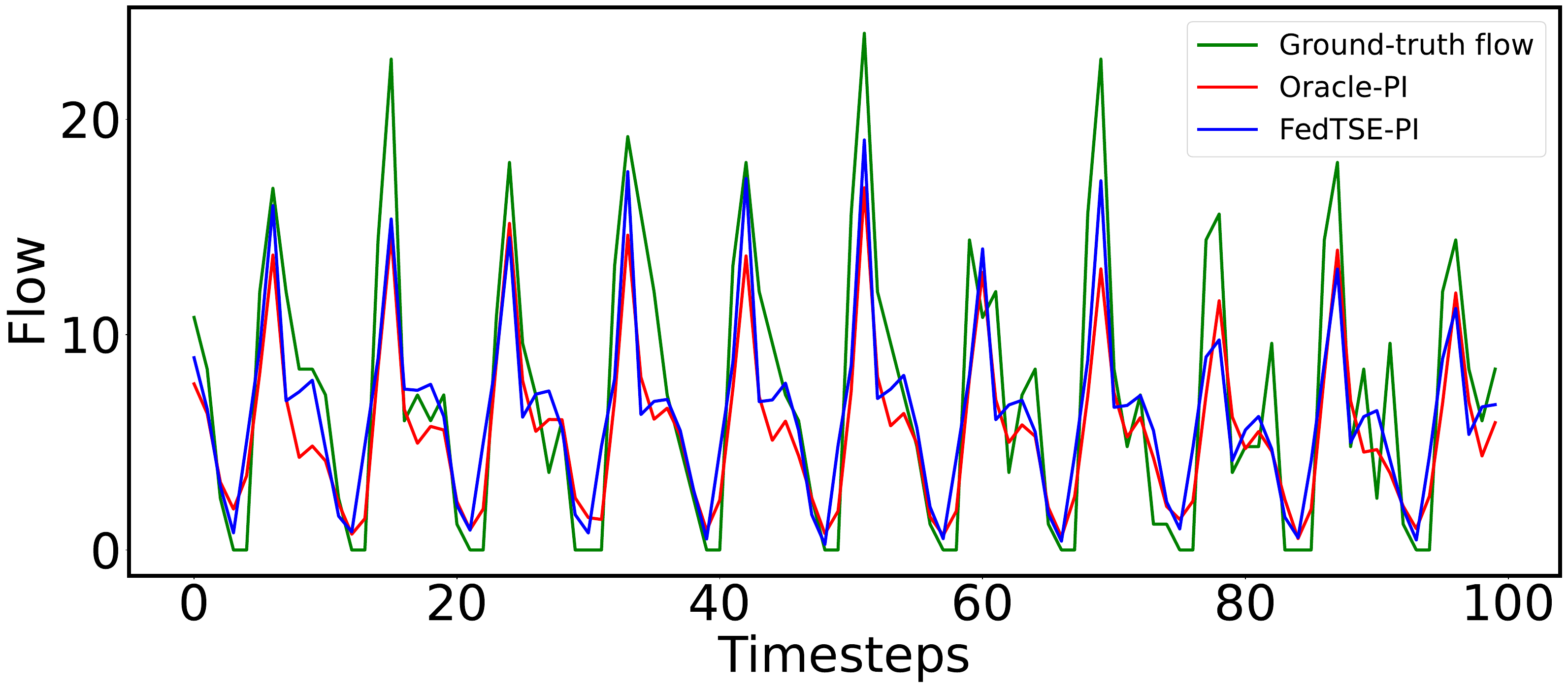}
         \caption{Flow estimation results of one link with no loop detector (veh/min/lane).}
         \label{fig:pi-est-2}
     \end{subfigure}
     \hfill
     \begin{subfigure}[h]{0.7\textwidth}
         \centering
         \includegraphics[width=\textwidth]{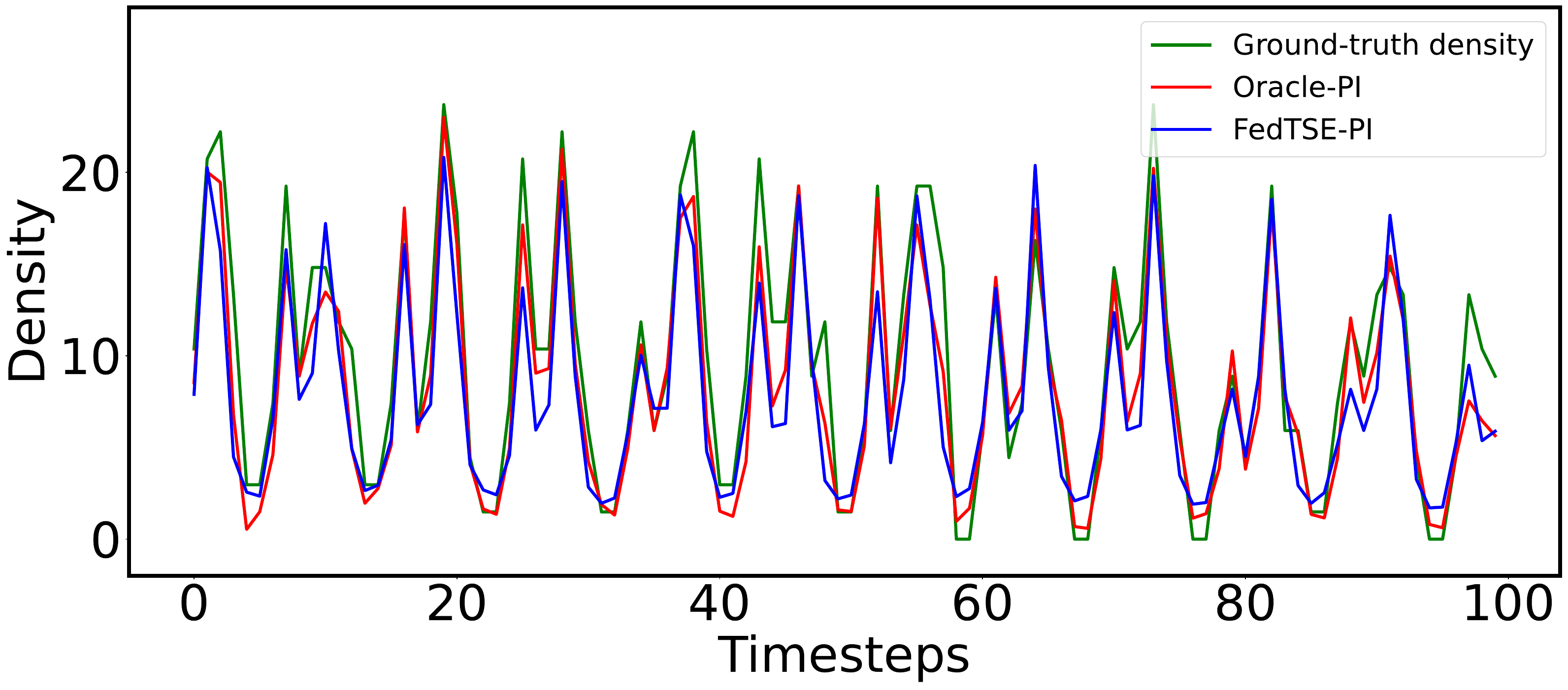}
         \caption{Density estimation results of one link with loop detector (veh/km/lane).}
         \label{fig:pi-est-3}
     \end{subfigure}
        \caption{Estimated density and flow comparison for FedTSE-PI. (20\% MP penetration rate)}
        \label{fig:pi-est}
\end{figure}

Moreover, let us take a closer look at the performance comparison of FedTSE-PI and Oracle-PI. Figure \ref{fig:pi-est} shows the estimated density and flow on links with and without MA observations, respectively, in scenarios with 20\% penetration rate. We can see that our proposed FedTSE-PI in blue line yields similar performance compared to Oracle-PI in red line. The vertical federated design in FedTSE-PI does not weaken model performance. For links equipped with loops, the density is well estimated (see Figure \ref{fig:pi-est-3}). This is because the flow can be perfectly estimated on these links thanks to the loop detectors. For links without loops, the estimated density can capture the main trend with some errors on peaks, which is because the information is not sufficient for these peaks due to low penetration rates. 

Overall, our proposed FedTSE-PI can yield desirable performance similar to Oracle-PI and significantly outperforms other state-of-the-art methods, which shows that FedTSE-PI achieves efficient tradeoff between privacy and estimation performance.
 
\noindent \textbf{Value of privacy protection in data fusion}. As mentioned above, MPs may be more incentivized to participate more actively in the data fusion and use higher-resolution data, which can in turn improve estimation accuracy. Here, we demonstrate this by comparing FedTSE-PI with FedTSE-PI-k and FedTSE-PI-v. As we can see that FedTSE-PI significantly outperforms FedTSE-PI-k (at least 14\% improvement) and FedTSE-PI-v (at least 60\% improvement), which demonstrate that the performance of FedTSE-PI can be improved by incorporating more measurements.  Similar conclusion can also be drawn on the benefits of using higher-resolution features by comparing FedTSE-PI with TSE-PI-p. 
This highlights the benefits of incorporating privacy-preserving mechanisms to data fusion, since this encourages MPs to share more measurements and features, and thus improve TSE performance.

\subsubsection{Sensitivity analysis on the number of loops}
Table \ref{tab-pi-loop} shows that more observations from MA could enhance the performance of FedTSE-PI. The RMSE of density and flow with 20\% MP penetration rate decreases more than 12\% and 17\% respectively from one loop to three loops. Although having only one observation, the density RMSE with 7.08 veh/km/lane is less than that of FedTSE-PI-k and UKF as shown in Table \ref{tab-pi}. This means that MP's data is crucial for MA when estimating traffic state. MA can still collaborate with MP to perform TSE with limited observations. Hence, it is necessary for MA to adopt a privacy-preserving framework to encourage MP's data-sharing.

\begin{table}[!htbp]
	\centering
	\caption{Sensitivity analysis on the number of loops}
	\label{tab-pi-loop}
		\begin{tabular}{lccccccccc}
    \hline
    & \multicolumn{1}{l}{Metrics} & \multicolumn{4}{c}{Density veh/km/lane} & \multicolumn{4}{c}{Flow veh/min/lane} \\
    \hline
    MP penetration rate (\%)& \multicolumn{1}{l}{} & 20 & 40 & 60 & 80 & 20 & 40 & 60 & 80 \\
    \hline
    \multirow{2}{*}{FedTSE-PI (three loops)} & RMSE & 6.20 & 5.49 & 4.99 & 4.52 & 3.36 & 2.85 & 2.65 & 2.48 \\
    & MAE  & 4.44 & 4.12 & 3.68 & 3.30 & 2.43 & 2.00 & 1.86 & 1.74 \\
    \multirow{2}{*}{FedTSE-PI (two loops)} & RMSE & 6.37 & 5.71 & 5.07 & 5.05 & 3.31 & 3.04 & 2.82 & 2.83 \\
    & MAE  & 4.81 & 4.23 & 3.81 & 3.72 & 2.39 & 2.18 & 1.98 & 2.02 \\
    \multirow{2}{*}{FedTSE-PI (one loop)} & RMSE & 7.08 & 6.06 & 5.63 & 5.26 & 4.07 & 3.63 & 3.11 & 2.86 \\
    & MAE  & 5.37 & 4.54 & 4.25 & 3.96 & 3.02 & 2.63 & 2.26 & 2.02\\
        \hline
\end{tabular}
\end{table}

\section{Conclusion} \label{sec-con}
In this paper, we propose two novel vertical FL-based methods, FedTSE and FedTSE-PI, for privacy-preserving learning-based data fusion of MA and MP's data for traffic state estimation. 
This is among the few pioneer works that address the cross-silo privacy concerns among ITS parties interested in collaborating and sharing heterogeneous datasets with different features. 
FedTSE enables multiple data owners to collaboratively train and apply a TSE model without having to explicitly exchange their private data. 
FedTSE-PI extends FedTSE to enhance the applicability in scenarios where ground-truth labels are not available, which is common for TSE due to the high cost of getting ground-truth traffic states. Case studies demonstrate that FedTSE and FedTSE-PI can preserve the privacy of data owners with minimum impact on the estimation performance, significantly outperforming the baselines where MPs only share partial or no data due to privacy concerns. Moreover, by protecting privacy, MPs can be more incentivized to participate more actively into data fusion and use higher-resolution data, which in turn enhances estimation accuracy. 

This work opens several promising future directions. First, the privacy guarantees of this work can be further enhanced by integrating with privacy-preserving data perturbation mechanisms such as differential privacy \citep{dwork2006differential} and statistical privacy filter \citep{nekouei2022model}. 
Second, we would like to relax the assumption that MPs are honest with no intention to forge data for their own benefit. This can be achieved by incorporating a cryptographic protocol based on Zero-Knowledge Proof \citep{fiege1987zero} similar to \cite{tsao2022trust}, which can detect the validity of the data used by MPs without MPs having to reveal their data. Third, we are interested in extending FedTSE-PI to consider 
the uncertainty quantification of both data and models, e.g., via Bayesian neural networks where the neural network weights are probabilistic distributions instead of deterministic values. Fourth, we would like to extend the vertical federated learning approach to perform traffic control by integrating it with reinforcement learning.

\printcredits

\section*{Acknowledgement}
The authors would like to thank Chaopeng Tan, Longhao Yan, and Jingyuan Zhou for fruitful discussions. This research was supported by the Singapore Ministry of Education (MOE) under its Academic Research Fund Tier 1 (A-8001183-00-00). This article solely reflects the opinions and conclusions of its authors and not Singapore MOE or any other entity.

\bibliographystyle{cas-model2-names}

\bibliography{cas-refs}


\end{document}